\documentclass{article}

 \usepackage[sglblindworkshop, final]{neurips_2025}

\usepackage[utf8]{inputenc} % allow utf-8 input
\usepackage[T1]{fontenc}    % use 8-bit T1 fonts
\usepackage{hyperref}       % hyperlinks
\usepackage{url}            % simple URL typesetting
\usepackage{booktabs}       % professional-quality tables
\usepackage{amsfonts}       % blackboard math symbols
\usepackage{nicefrac}       % compact symbols for 1/2, etc.
\usepackage{microtype}      % microtypography
\usepackage{xcolor}         % colors

\usepackage{algorithm}
\usepackage{algpseudocode}
\usepackage{amsmath}
\usepackage{amsthm}
\usepackage{amsfonts}
\usepackage{tabularx}
\usepackage{amssymb}

\workshoptitle{Foundations of Reasoning in Language Models (FoRLM)}
\title{\texttt{Monitor-Generate-Verify (MGV)}:\\Formalising Metacognitive Theory for\\Language Model Reasoning}

\author{%
Nick Oh \\
socius labs\\ 
London, UK\\
\texttt{nick.sh.oh@socius.org}\\
\And
Fernand Gobet \\
Centre for Philosophy of \\
Natural and Social Science (CPNSS)\\
London School of Economics\\
London, UK\\
\texttt{f.gobet@lse.ac.uk}
}

\begin{document}

\maketitle

\begin{abstract}
Test-time reasoning architectures such as those following the \textit{Generate-Verify} paradigm, where a model iteratively refines or verifies its own generated outputs, prioritise generation and verification but exclude the monitoring processes that determine when and how reasoning should begin. This omission may contribute to the prefix dominance trap, in which models commit early to suboptimal reasoning paths and seldom recover, yielding roughly 20\% accuracy loss. We address this architectural gap by proposing the Monitor-Generate-Verify (MGV) framework, a computational translation of Flavell's and Nelson and Narens' metacognitive theories that preserves their psychological detail. MGV extends the Generate-Verify paradigm by adding explicit monitoring that captures metacognitive experiences (from difficulty assessments to confidence judgements) before generation begins and refines future monitoring through verification feedback. Though we present no empirical validation, MGV provides a vocabulary for diagnosing component-level failures in reasoning systems, suggests specific architectural interventions for future designs, and identifies connections to resource-rational analysis that may ground its mechanisms in normative principles.
\end{abstract}

\section{Introduction}

Once language models commit to an initial reasoning strategy, subsequent verification rarely helps; this \textit{prefix dominance trap} causes nearly 20\% performance degradation when models choose suboptimal approaches, with limited recovery possible through refinement \citep{luo2025learning}. Today's dominant Generate-Verify test-time reasoning architectures \citep{weng2023large, madaan2023self, lee2025revise, zhang2024small} exemplify this limitation through their very design. They operate through immediate generation followed by iterative refinement, without assessing task characteristics or selecting appropriate strategies before generating solutions. What architectural vocabulary might help diagnose what these systems lack?

Cognitive psychology offers a candidate answer. Before attempting complex tasks, humans engage in metacognitive monitoring, assessing difficulty, retrieving relevant strategies, and establishing confidence criteria, often without conscious deliberation \citep{flavell1979metacognition, nelson1990metamemory}. This monitoring capacity transforms uncertainty into tractable signals that guide subsequent action. \citeauthor{flavell1979metacognition}'s model specifies how metacognitive knowledge and experience interact to enable strategy selection, while \citeauthor{nelson1990metamemory}' metamemory framework details how confidence thresholds and adaptive search mechanisms emerge from hierarchical monitoring and control. These theories suggest that the Generate-Verify paradigm may be missing an entire phase of processing that precedes generation.

However, translating these psychological insights into architectural recommendations requires computational specification. Resource-rational analysis has made significant progress here, with \citet{callaway2024optimal} formalising \citeauthor{nelson1990metamemory}' metamemory theory as a meta-level MDP that derives optimal stopping policies from cost-benefit principles. While their approach offers normative grounding and empirical validation, the present work pursues a complementary goal: instead of deriving optimal behaviour from computational constraints, we translate the detailed psychological vocabulary of \citet{flavell1979metacognition} and \citet{nelson1990metamemory} into algorithmic form, preserving structure that resource-rational approaches may abstract away.

The resulting Monitor-Generate-Verify (MGV) framework extends the Generate-Verify paradigm by adding explicit monitoring that captures metacognitive experiences before generation begins and refines future monitoring through verification feedback. Our formalisation preserves constructs including the tripartite organisation of metacognitive knowledge, the distinction between difficulty and evaluative experiences, dual-counter evidence accumulation for feeling-of-knowing, and satisficing threshold dynamics. We offer no empirical validation and no normative justification for these architectural choices. The contribution is a detailed computational translation that provides diagnostic vocabulary for understanding what reasoning systems might lack, suggesting where architectural interventions might prove productive.

The remainder of this paper is organised as follows. Section~\ref{sec:related} situates our contribution within resource-rational approaches to metacognition and LLMs. Section~\ref{sec:frameworks} presents algorithmic formalisations of Flavell's and Nelson and Narens' theories, translating psychological concepts into computational structures. Section~\ref{sec:limit} acknowledges limitations and outlines future directions.

\section{Related Work}
\label{sec:related}

This section situates the present work within two related research programmes. The first is resource-rational analysis, a cognitive modelling paradigm that formalises metacognition as cost-benefit optimisation over computational resources. The second is recent work applying these principles to language models. We review each in turn before clarifying how the present work differs in its foundational commitments.

\subsection{Resource-Rational Analysis of Metacognition}

Resource-rational analysis is a cognitive modelling paradigm that reconceives bounded rationality not as a departure from optimality but as optimality under constraints \citep{griffiths2015rational, griffiths2019doing, lieder2020resource}. Making good decisions requires computation, yet computation itself is costly. This tension between the value of deliberation and its cost gives rise to a class of meta-problems: problems not about what to decide, but about how to decide. \textit{How much should an agent think before acting}? \textit{Which cognitive strategy should be deployed for a given problem}? \textit{Which aspects of a decision path merit exploration, and which can be safely ignored}? These questions concern the allocation of computational resources, and answering them requires reasoning about reasoning itself.

Resource-rational analysis provides a framework for addressing such meta-problems by treating cognitive constraints as part of the optimisation problem rather than as departures from rationality. Under metalevel rationality, an agent is assessed not by the expected utility of actions taken, but by how well its cognitive algorithm trades off decision quality against deliberation costs. From this perspective, rationality concerns not only making good decisions but also employing efficient cognitive strategies. The framework reduces meta-reasoning\footnote{Meta-reasoning should be distinguished from meta-learning. Meta-reasoning concerns how to allocate limited computational resources during cognition, deciding which computations to perform and when to stop deliberating. Meta-learning concerns how to structure the learning process itself, exploiting knowledge about the statistical structure of learning environments to make better use of limited data. Both address resource limitations, but at different levels: meta-reasoning optimises the use of time and cognitive effort within a task, whereas meta-learning optimises the acquisition of knowledge across tasks. For a treatment that situates both perspectives within the resource-rational framework, see \citet{griffiths2019doing}.} to a decision-theoretic problem: \textit{for any computation the agent could perform, there exists a value of computation (VOC) that quantifies the expected improvement in decision quality minus the cost of deliberation}.

This cognitive modelling paradigm provides a common structure that different research programmes have instantiated in distinct ways, each addressing a different facet of the meta-reasoning problem. \citet{lieder2017strategy} modelled \textit{strategy selection}: rather than asking whether to think more, this formulation asks how to think, treating the choice among cognitive strategies as a contextual bandit in which the agent learns to predict which mode of reasoning best suits each type of problem (Appendix~\ref{app:strategy}). \citet{lieder2018rational} modelled \textit{cognitive control}: by casting control specification as a belief-MDP and introducing a learned approximation, this formulation explains how control exhibits plasticity, adapting through experience to select signals that optimally trade off effort against goal achievement (Appendix~\ref{app:control}). \citet{callaway2022rational} modelled \textit{planning}: by embedding deliberation within a meta-level MDP, this formulation captures the sequential structure of mental simulation, recognising that the value of a computation depends not only on its immediate effect but also on the future computations it enables (Appendix~\ref{app:planning}). \citet{callaway2024optimal} modelled \textit{memory recall}: by treating retrieval as evidence accumulation supervised by an optimal stopping process, this formulation reveals the adaptive function of feeling-of-knowing, providing the monitoring signal that enables efficient termination of searches unlikely to succeed while sustaining effort when retrieval remains probable (Appendix~\ref{app:recall}).

These four instantiations span a range of formal structures, from contextual bandits to belief-MDPs, and a range of cognitive phenomena, from high-level strategy choice to low-level memory retrieval. Yet they share a common theoretical commitment: the mind's computational processes are themselves subject to rational optimisation. Appendix~\ref{app:resource-rational} provides a detailed technical treatment of these frameworks, comparing their formal structures and identifying shared algorithmic motifs. Here, we turn to recent work that has begun extending these principles beyond cognitive modelling to the training of language models.

\subsection{Rational Metareasoning for Language Models}

While the research reviewed above models human cognition, recent work has begun applying the VOC framework directly to language model training. \citet{desabbata2024rational} introduced RaM (Rational Metareasoning), which trains LLMs to use chain-of-thought reasoning selectively based on task difficulty. Their approach defines the reward of a reasoning chain $z$ as the difference between its utility and cost:
\begin{equation*}
R_\pi(x, y, z) = U_\pi(z | x, y) - C(z),
\end{equation*}
where utility $U_\pi(z | x, y) = \log \pi_\theta(y | z, x) - \log \pi_\theta(y | x)$ quantifies the increase in likelihood of generating the correct answer $y$ when the chain of thought $z$ is included, and cost $C(z) = \gamma \cdot l(z)$ is proportional to the number of tokens. This formulation mirrors the equation, with reasoning tokens corresponding to intermediate computations and model outputs corresponding to actions. Models trained with this objective learn to generate shorter reasoning chains for easier problems and longer chains for harder problems, demonstrating that VOC-based training can induce adaptive computation in neural systems. This work provides empirical evidence that resource-rational principles can be operationalised in modern architectures, though it focuses on optimising existing reasoning capabilities rather than introducing new metacognitive components.

\subsection{Relationship to the Present Work}

Formalising metacognitive theory requires choosing what to treat as primitive and what to derive. Resource-rational analysis treats computational operations as primitive and derives metacognitive behaviour from cost-benefit optimisation. The present work makes the opposite choice, treating psychological constructs as primitive and translating them directly into algorithmic form. Put simply, instead of deriving that people think the way they do because it is efficient given limited cognitive resources, this work treats psychological ideas as the foundation and directly formalises them in computational terms.

This difference in approach leads to different contributions. Resource-rational analysis provides normative grounding, explaining why certain metacognitive policies are optimal given computational constraints. The present work provides psychological granularity, preserving detailed structure from the source theories that resource-rational models abstract away. Section~\ref{sec:frameworks} makes this concrete by formalising constructs including tripartite metacognitive knowledge, dual-counter evidence accumulation, and satisficing threshold dynamics. Section~\ref{sec:limit} returns to the relationship between these approaches and examines how they might complement one another.

\section{Monitor-Generate-Verify (MGV)}
\label{sec:frameworks}

\citet{flavell1979metacognition} and \citet{nelson1990metamemory} developed seminal theories of how metacognition coordinates cognitive processes through monitoring and control loops. These frameworks, though developed for human cognition, offer potential blueprints for computational systems. \citeauthor{flavell1979metacognition}'s model provides a dynamic architecture where metacognitive knowledge and experience guide strategy selection and verification, while \citeauthor{nelson1990metamemory}' metamemory framework specifies how confidence thresholds and adaptive search mechanisms emerge from hierarchical monitoring and control. By computationally formalising these psychological theories, we establish Monitor-Generate-Verify (MGV) as a theoretical framework for understanding how explicit metacognitive mechanisms could address the architectural limitations of current reasoning systems. The following subsections present detailed formalisations that translate these cognitive psychology insights into algorithmic structures, revealing both what current architectures lack and how metacognitive principles might be operationalised computationally.

\subsection{Flavell's Model of Metacognition}

\citet{flavell1979metacognition} conceptualises metacognition as a dynamic control architecture comprising four interacting components: \textit{metacognitive knowledge}, \textit{metacognitive experience}, \textit{goals} (or tasks), and \textit{actions} (or strategies). Rather than operating as independent modules, these components form an integrated system characterised by continuous bidirectional influences, positioning metacognition as a self-regulating system capable of adaptive control over cognitive processes. We present the core computational structure below, with a complete mathematical formalisation provided in Appendix~\ref{app:flavell}.

\subsubsection{Cognitive Monitoring}

The regulation process begins with initialisation, where task $\mathcal{T}$ and goal $\mathcal{G}$ establish the initial state $\mathcal{S}_0 = f(\mathcal{T}, \mathcal{G})$. While \citet{flavell1979metacognition} treats goals and tasks as equivalent, we maintain a computational distinction. $\mathcal{T}$ represents the cognitive enterprise while $\mathcal{G}$ specifies success criteria, enabling clearer analysis of metacognitive processes.

\begin{algorithm}[h]
\caption{Flavell's Model of Cognitive Control}
\begin{algorithmic}[1]
\State \textbf{Initialise:} $\mathcal{S}_0 \gets f(\mathcal{T}, \mathcal{G})$; $\tau \gets 0$
\While{$\mathcal{S}_\tau = \text{ACTIVE}$}
    \State \textbf{// MONITOR: Retrieve knowledge \& assess experience}
    \State $\mathcal{MK}_\tau \gets$ \textbf{if} $\tau = 0$ \textbf{then} retrieve($\mathcal{MK}, \mathcal{T}, \mathcal{G}$) 
    \State \hspace{2.5em} \textbf{else} $\mathcal{MK}_{\tau-1} \cup$ retrieve($\mathcal{MK}, \mathcal{ME}_{\tau-1}$)
    \State $\mathcal{ME}_{\text{difficulty}}^\tau \gets$ feel($\mathcal{T}$, Outcomes$_{\tau-1}$) $\oplus$ assess($\mathcal{T}$, $\mathcal{MK}_\tau$)
    \State \textbf{// GENERATE: Select \& execute cognitive strategy}
    \State $\mathcal{CS}_\tau \gets$ select($s \in \mathcal{MK}_{\text{Strategy}} \mid \mathcal{ME}_{\text{difficulty}}^\tau, \mathcal{MK}_\tau, \mathcal{T}, \mathcal{G}$)
    \State $\mathcal{CO}_\tau \gets$ execute($\mathcal{CS}_\tau, \mathcal{T}, \mathcal{G}$)
    \State \textbf{// VERIFY: Evaluate progress \& update knowledge}
    \State $\mathcal{ME}_{\text{evaluative}}^\tau \gets$ assess($\mathcal{CO}_\tau, \mathcal{MK}_\tau$)
    \State $\mathcal{MS}_\tau \gets$ select($s \in \mathcal{MK}_{\text{Strategy}}^{\text{meta}} \mid \mathcal{ME}_{\text{evaluative}}^\tau$)
    \State $\mathcal{MO}_\tau \gets$ execute($\mathcal{MS}_\tau, \mathcal{CO}_\tau, \mathcal{MK}_\tau, \mathcal{G}$)
    \State $\mathcal{MK} \gets$ update($\mathcal{MK}$, $\Phi_\tau$) where $\Phi_\tau = (\mathcal{ME}_\tau, \text{Strategy}_\tau, \text{Outcome}_\tau)$
    \State $\mathcal{S}_{\tau+1} \gets$ \textbf{if} goal\_achieved($\mathcal{CO}_\tau, \mathcal{G}$) \textbf{then} TERMINATE \textbf{else} ACTIVE
    \State $\tau \gets \tau + 1$
\EndWhile
\end{algorithmic}
\end{algorithm}

The \textbf{monitoring} phase activates metacognitive knowledge differently across cycles. Initial cycles rely solely on task-goal combinations, while subsequent cycles incorporate emerging metacognitive experiences from $\tau-1$ that trigger additional relevant knowledge. According to \citet{flavell1979metacognition}, this knowledge comprises three categories: \textit{agent variables} ($\mathcal{MK}_{\text{Agent}}$) representing learned self-models of performance patterns and processing preferences; \textit{task variables} ($\mathcal{MK}_{\text{Task}}$) capturing knowledge about cognitive situation assessment including information characteristics and task demands; and \textit{strategy variables} ($\mathcal{MK}_{\text{Strategy}}$) encompassing knowledge about the effectiveness of both cognitive strategies (problem-solving procedures) and metacognitive strategies (monitoring and regulation processes). These categories function as an integrated system where task variables diagnose cognitive demands, strategy variables prescribe responses, and agent variables contextualise both within the agent's capabilities.

The monitoring phase also generates metacognitive experiences of difficulty ($\mathcal{ME}_{\text{difficulty}}^\tau$), which \citet[p.~909]{flavell1979metacognition} describes as subjective feeling-of-complexity, comprehension challenges, or sensing that material exceeds current capabilities. These experiences evolve through iterative assessments, progressing from initial coarse feelings to increasingly nuanced evaluations of specific challenge sources.

During the \textbf{generation} phase, metacognitive experiences function as computational signals that require interpretation through metacognitive knowledge to guide strategy selection. The process follows a two-phase pattern. First, $\mathcal{MK}_{\text{Strategy}}$ transforms general difficulty \textit{signals} into precise diagnostic patterns (e.g., ``content uncertainty with unknown terms'' or ``procedural confusion from missing steps''). Second, these refined patterns activate corresponding cognitive strategies. The selected strategy $\mathcal{CS}_\tau$ is then executed to produce cognitive outcomes $\mathcal{CO}_\tau$, generating feedback that provides both task progress information and context for subsequent monitoring.

The \textbf{verification} phase evaluates these outcomes, triggering what \citet[p.~909]{flavell1979metacognition} describes as additional metacognitive experiences about performance rather than difficulty. These evaluative experiences ($\mathcal{ME}_{\text{evaluative}}^\tau$) activate metacognitive strategies that assess whether outcomes form a coherent whole, appear plausible and consistent with prior knowledge, and provide an avenue to the goal. The specific metacognitive strategy $\mathcal{MS}_\tau$ selected depends on the nature of the evaluative signal: uncertainty about validity triggers plausibility checking, sensing incompleteness activates coherence assessment, and so forth. Notably, these experiences can add to, delete from, or revise the metacognitive knowledge base through Piagetian mechanisms \citep{flavell1963developmental}, with the complete experience tuple $\Phi_\tau$ updating $\mathcal{MK}$ for future cycles.

\subsubsection{Memory and Learning Gaps}

A significant limitation in \citeauthor{flavell1979metacognition}'s model is the absence of explicit working memory mechanisms for storing information across monitoring cycles. The model does not specify where $\mathcal{ME}_{\text{difficulty}}^\tau$ resides during strategy execution, how $\mathcal{CO}_\tau$ is maintained during evaluation, or how experience patterns across cycles are retained for subsequent processing. This absence precludes sophisticated termination criteria that would require access to historical monitoring data across the complete sequence $\Phi = (\Phi_0, \Phi_1, \ldots, \Phi_T)$.

With an explicit memory component, the model could implement comprehensive abandonment criteria that evaluate: (1) repeated strategy failures indicated by consistently negative $\mathcal{MO}_\tau$ across multiple cycles, suggesting task intractability; (2) resource constraints where cumulative effort across $\Phi_0$ to $\Phi_\tau$ exceeds acceptable limits relative to $\mathcal{MK}_{\text{Agent}}$; (3) goal displacement where evolving $\mathcal{ME}_{\text{evaluative}}^\tau$ signals that alternative objectives have become more salient than the original $\mathcal{G}$; and (4) insurmountable goal-state discrepancy where the pattern of $\mathcal{CO}_\tau$ outcomes reveals fundamental incompatibility with $\mathcal{G}$ achievement.

A related temporal limitation concerns metacognitive knowledge acquisition and refinement. While \citeauthor{flavell1979metacognition} acknowledges that experiences can `add to', `delete from', or `revise' the knowledge base, the model assumes pre-existing $\mathcal{MK}$ without specifying learning mechanisms -- how unsuccessful strategies refine strategy knowledge, or how repeated encounters improve task assessments.

Such memory-dependent termination decisions and learning-dependent knowledge refinement would better reflect real-world metacognitive monitoring, where individuals track cumulative progress patterns and recognise when persistence becomes counterproductive, while simultaneously refining their metacognitive knowledge through experience. These limitations point towards the necessity for more sophisticated architectural frameworks that explicitly model the temporal dynamics of metacognitive information storage and retrieval as well as the acquisition and refinement of metacognitive knowledge -- considerations that become central to \citeauthor{nelson1990metamemory}' metamemory architecture. 

\subsection{Nelson and Narens' Model of Metamemory}

\citet{nelson1990metamemory} establish metacognition as fundamentally hierarchical, distinguishing between \textit{object-level} and \textit{meta-level} processes. The \textit{object-level} comprises the cognitive operations themselves, such as searching memory for a target item. The \textit{meta-level} supervises these operations, by monitoring progress toward the goal (e.g., how close is recall?) and controlling how long the process is allowed to continue (e.g., should the search persist or terminate?). This supervision relies on two distinct information flows: monitoring conveys information upward from object-level to meta-level, while control conveys decisions downward from meta-level to object-level. These relationships are logically independent and asymmetric: the meta-level maintains a model of the object-level, while the object-level operates without corresponding meta-level representation. We present the core computational structure below, with complete mathematical formalisation provided in Appendix~\ref{app:nelson-narens}.

\subsubsection{Acquisition Process}

The acquisition process begins with establishing the \textit{norm of study} $\mathcal{N}_s = \rho^* \times (1 + \delta_{\text{retention}})$, where $\rho^*$ represents target performance and $\delta_{\text{retention}}$ captures beliefs about memory decay over interval $\tau_{\text{delay}}$. This operationalises abstract goals into quantified mastery criteria that anticipate forgetting. Following \citet{ericsson1984protocol}, monitoring occurs within working memory (STM), with information from long-term memory (LTM) accessed probabilistically via $\text{retrieve}_\theta(\cdot)$ where $\theta$ represents access probability \citep{atkinson1968human}.

\begin{algorithm}[h]
\caption{Nelson and Narens' Model of Acquisition}
\begin{algorithmic}[1]
\State \textbf{Initialise:} $\mathcal{MK}_0^{\text{STM}} \gets$ retrieve$_\theta(\mathcal{MK}, \mathcal{T}, \mathcal{G})$
\State $\mathcal{N}_s \gets \rho^* \times (1 + $ formulate$(\mathcal{MK}_0^{\text{STM}}, \tau_{\text{delay}}, \mathcal{T}, \mathcal{G}))$
\State $\mathcal{J}_0 \gets \{1, \ldots, N\}$; $\tau \gets 1$; $\Phi_0^{\text{STM}} \gets \emptyset$
\While{$\mathcal{J}_\tau \neq \emptyset$}
    \State \textbf{// MONITOR: Assess mastery via EOL/FOK}
    \State $\mathcal{MK}_\tau^{\text{STM}} \gets \mathcal{MK}_{\tau-1}^{\text{STM}} \cup$ retrieve$_\theta(\mathcal{MK}, \mathcal{ME}_{\tau-1})$
    \For{each $j \in \mathcal{J}_\tau$}
        \State $\mathcal{ME}_{\tau,j}[1] \gets$ \textbf{if} $\tau = 1$ \textbf{then} EOL($i_j$) \textbf{else} FOK($i_j, \mathcal{CO}_{\tau-1,j}$)
    \EndFor
    \State \textbf{// GENERATE: Allocate resources \& select strategies}
    \For{each $j \in \mathcal{J}_\tau$}
        \State $r_{\tau,j} \gets R_{\text{total}} \times (\mathcal{ME}_{\tau,j}[1])^{-1} / \sum_k (\mathcal{ME}_{\tau,k}[1])^{-1}$
        \State $\sigma_{\tau,j} \gets$ select$(s \in \mathcal{MK}_{\text{Strategy}} \mid i_j, r_{\tau,j}, \mathcal{ME}_{\tau,j})$
        \State $\mathcal{CO}_{\tau,j} \gets$ execute$(i_j, r_{\tau,j}, \sigma_{\tau,j})$
    \EndFor
    \State \textbf{// VERIFY: Judge learning \& update items}
    \For{each $j \in \mathcal{J}_\tau$}
        \State JOL$_{\tau,j} \gets$ feel$(i_j, \mathcal{CO}_{\tau,j}) \oplus$ assess$(i_j, \mathcal{CO}_{\tau,j}, \mathcal{MK}_\tau^{\text{STM}})$
        \State $\mathcal{ME}_{\tau,j}[2] \gets$ JOL$_{\tau,j}$
        \State $\Phi_{\tau}^{\text{STM}} \gets \Phi_{\tau}^{\text{STM}} \cup \{(\mathcal{ME}_{\tau,j}, i_j, r_{\tau,j}, \sigma_{\tau,j}, \mathcal{CO}_{\tau,j})\}$
    \EndFor
    \State $\mathcal{J}_{\tau+1} \gets \{j \in \mathcal{J}_\tau : \mathcal{N}_s - \text{JOL}_{\tau,j} > 0\}$; $\tau \gets \tau + 1$
\EndWhile
\State $\mathcal{MK} \gets$ consolidate$_\psi(\mathcal{MK}, \Phi_{\tau}^{\text{STM}})$ \Comment{Experience to LTM}
\end{algorithmic}
\end{algorithm}

The \textbf{monitoring} phase generates metacognitive experiences as multidimensional vectors. Ease-of-learning (EOL) provides initial difficulty assessment, while feeling-of-knowing (FOK) incorporates prior outcomes to refine mastery judgements. These phenomenological experiences serve as primary input for control decisions \citep[p.~160]{nelson1990metamemory}. During the \textbf{generation} phase, resource allocation operates inversely to EOL/FOK values -- items with lower metacognitive confidence receive proportionally more resources $r_{\tau,j} = R_{\text{total}} \times w_j/\sum_k w_k$ where $w_j = (\mathcal{ME}_{\tau,j}[1])^{-1}$. Strategy selection integrates these metacognitive inputs to map appropriate learning methods to individual items.

The \textbf{verification} phase employs judgement-of-learning (JOL) to evaluate mastery following cognitive outcomes. Items achieving the norm of study ($\text{JOL}_{\tau,j} \geq \mathcal{N}_s$) are removed from further consideration, while those below threshold remain in $\mathcal{J}_{\tau+1}$ for continued learning. The complete experience tuple accumulates in working memory as $\Phi_{\tau}^{\text{STM}}$, subsequently undergoing consolidation to LTM at encoding rate $\psi$.

\subsubsection{Retrieval Process}

The retrieval process implements \citeauthor{nelson1990metamemory}' dual-counter FOK hypothesis, where $\text{FOK}^{+}$ accumulates evidence for information presence whilst $\text{FOK}^{-}$ accumulates evidence for absence, consistent with `knowing not' \citep{kolers1976knowing}. Initial thresholds are personalised through metacognitive calibration history: $\lambda_{\text{FOK}}^{(0)} = \text{median}(\{||\text{FOK}|| : \text{successful retrievals in } \mathcal{MK}_0^{\text{STM}}\})$ and $\lambda_{\text{confidence}}^{(0)} = \text{median}(\{\text{confidence} : \text{correct outputs in } \mathcal{MK}_0^{\text{STM}}\})$, embodying the No-Magic Hypothesis by utilising recallable metacognitive knowledge.

\begin{algorithm}[h]
\caption{Nelson and Narens Model of Retrieval}
\begin{algorithmic}[1]
\State \textbf{Initialise:} $\mathcal{MK}_0^{\text{STM}} \gets$ retrieve$_\theta(\mathcal{MK}, \mathcal{Q})$
\State $\lambda_{\text{FOK}}^{(0)}, \lambda_{\text{confidence}}^{(0)} \gets$ calibrate$(\mathcal{MK}_0^{\text{STM}})$; $\tau \gets 0$; $\Omega_0^{\text{STM}} \gets \emptyset$
\While{search active}
    \State \textbf{// MONITOR: Assess dual-counter FOK}
    \State $\mathcal{MK}_\tau^{\text{STM}} \gets \mathcal{MK}_{\tau-1}^{\text{STM}} \cup$ retrieve$_\theta(\mathcal{MK}, \text{FOK}_{\tau-1})$ if $\tau > 0$
    \State $[\text{FOK}_\tau^{+}, \text{FOK}_\tau^{-}] \gets$ feel$(\mathcal{Q}, \mathcal{A}_{\tau-1}) \oplus$ assess$(\mathcal{Q}, \mathcal{A}_{\tau-1}, \mathcal{MK}_\tau^{\text{STM}})$
    \State \textbf{// Determine search intensity based on FOK evidence}
    \If{$||\text{FOK}_\tau|| < \lambda_{\text{FOK}}^{(\tau)}$} 
        \State $\mathcal{S}_\tau \gets \text{ACTIVE}_{\text{intensive}}$ \Comment{Insufficient evidence}
    \ElsIf{$\text{FOK}_\tau^{+} > \text{FOK}_\tau^{-}$}
        \State $\mathcal{S}_\tau \gets \text{ACTIVE}_{\text{standard}}$ \Comment{Positive dominance}
    \Else
        \State \textbf{break} \Comment{Negative dominance: terminate}
    \EndIf
    \State \textbf{// GENERATE: Attend to cues \& automatic search}
    \State cue$_\tau \gets$ attend$_{[\text{intensive/standard}]}(\mathcal{Q}, \mathcal{MK}_\tau^{\text{STM}})$ based on $\mathcal{S}_\tau$
    \State $\mathcal{A}_\tau \gets$ search$_{\text{auto}}($cue$_\tau)$ \Comment{Automatic pattern recognition}
    \State \textbf{// VERIFY: Evaluate answer \& adjust thresholds}
    \State confidence$_\tau \gets$ assess$(\mathcal{A}_\tau, \mathcal{Q}, \mathcal{MK}_\tau^{\text{STM}})$ if $\mathcal{A}_\tau \neq$ null
    \If{$\mathcal{A}_\tau \neq$ null $\wedge$ confidence$_\tau \geq \lambda_{\text{confidence}}^{(\tau)}$}
        \State \textbf{output} $\mathcal{A}_\tau$; \textbf{break}
    \ElsIf{$\mathcal{A}_\tau =$ null $\wedge \text{FOK}_\tau^{-} > \text{FOK}_\tau^{+}$}
        \State \textbf{output} null; \textbf{break} \Comment{Omission}
    \EndIf
    \State $\Omega_\tau^{\text{STM}} \gets \Omega_\tau^{\text{STM}} \cup \{(\text{FOK}_\tau, \text{cue}_\tau, \mathcal{A}_\tau, \text{confidence}_\tau)\}$
    \State $\beta_\tau \gets \exp(-\alpha \cdot (\tau + |\{$failed attempts in $\Omega_\tau^{\text{STM}}\}|))$
    \State $\lambda_{\text{confidence}}^{(\tau+1)}, \lambda_{\text{FOK}}^{(\tau+1)} \gets \lambda^{(0)} \cdot \beta_\tau$ \Comment{Satisficing}
    \State $\tau \gets \tau + 1$
\EndWhile
\State $\mathcal{MK} \gets$ consolidate$_\psi(\mathcal{MK}, \Omega_\tau^{\text{STM}})$ \Comment{Experience to LTM}
\end{algorithmic}
\end{algorithm}

The \textbf{monitoring} phase employs rapid FOK assessment that operates faster than actual recall \citep{reder1987strategy}, enabling efficient search control. When FOK magnitude falls below threshold ($||\text{FOK}_\tau|| < \lambda_{\text{FOK}}^{(\tau)}$), insufficient evidence triggers intensive cue attention to gather additional metacognitive information. With sufficient evidence, positive dominance ($\text{FOK}_\tau^{+} > \text{FOK}_\tau^{-}$) warrants continued search, while negative dominance justifies termination.

The \textbf{generation} phase reflects \citeauthor{nelson1990metamemory}' insight that search execution is automatic once initiated -- \textit{conscious control operates through cue attention intensity rather than strategy selection}. The automatic search process $\text{search}_{\text{auto}}(\text{cue}_\tau)$ operates through pattern recognition, potentially yielding identical results across consecutive cycles due to its deterministic nature.

\textbf{Verification} distinguishes two error pathways: commission errors (outputting incorrect answers with high confidence) and omission errors (terminating without answers following prolonged search). Following satisficing principles \citep{simon1979models}, both confidence and FOK thresholds undergo dynamic adjustment: $\lambda^{(\tau+1)} = \lambda^{(0)} \cdot \beta_\tau$ where $\beta_\tau = \exp(-\alpha \cdot \text{burden})$ captures accumulating search costs. This ensures previously inadequate answers may become acceptable as search burden increases, preventing exhaustive search behaviour.

\subsubsection{Memory Consolidation and Knowledge Evolution}

A distinctive strength of \citeauthor{nelson1990metamemory}' framework lies in its explicit treatment of long-term memory (LTM) as both a repository and an evolving knowledge base. During acquisition and retrieval, the experience tuples accumulated in working memory ($\Omega_T^{\text{STM}}$) undergoes consolidation into LTM at encoding rate $\psi$:

While \citeauthor{nelson1990metamemory} do not explicitly specify the timing of this consolidation process, it likely occurs during the verification stage at rate $\psi$, potentially operating below conscious awareness. This consolidation mechanism enables the global metacognitive knowledge base to evolve through accumulated experience, distinguishing \citeauthor{nelson1990metamemory}' approach from more static metacognitive frameworks. The probabilistic retrieval function $\text{retrieve}_\theta(\mathcal{MK}, \cdot)$ subsequently accesses this enriched knowledge base, creating a dynamic feedback loop where metacognitive experiences inform future metacognitive assessments.

\section{Limitations and Future Work}
\label{sec:limit}

\subsection{From Specification to Implementation}

The MGV framework specifies what should be computed but not how to compute it in neural architectures. Operationalising constructs like $\mathcal{ME}_{\text{difficulty}}$ or dual-counter FOK requires identifying appropriate neural correlates or designing explicit computational mechanisms. For language models, this might involve obtaining logit-based confidence ratings \citep{kumaran2025overconfidence}, using entropy-based proxies for metacognitive experiences, or implementing explicit evidence accumulators. Recent work suggests that LLMs access metacognitive signals they cannot adequately articulate, with implicit confidence measures outperforming explicit verbalisation \citep{wang2025decoupling, tian2023just, xiong2023can, cash2024quantifying, griot2025large, lindsey2025biology}. Neural analysis reveals a lower-dimensional ``metacognitive space'' where monitoring signals correspond to linearly separable directions \citep{ji2025language, zou2023representation, liu2023aligning}, consistent with cognitive findings that metacognition operates on abstracted representations \citep{reder1987strategy}.

\subsection{Scope of Formalisation}

The present work formalises two foundational theories, but metacognition research extends well beyond Flavell and Nelson and Narens. A more complete computational account of metacognition should integrate subsequent theoretical developments. Additionally, our formalisations make interpretive choices where the source theories are ambiguous or silent. For instance, the exponential decay of satisficing thresholds reflects our reading of the theories rather than explicit specifications in the original texts. Alternative formalisations are possible and might yield different architectural implications.

\subsection{Normative Grounding}

The framework lacks principled justification for its architectural choices. Resource-rational analysis explains why certain metacognitive policies are optimal given computational constraints. MGV, by contrast, specifies mechanisms without explaining why those mechanisms are appropriate. As \citet[pp.~17--18]{callaway2024optimal} noted regarding related work in metamemory, this paper shares the limits of \citet{metcalfe1993novelty}, \citet{koriat1996monitoring}, and \citet{bennett2017bayesian} in focusing on functional aspects without taking an explicitly optimal approach. That is, while resource-rational analysis could potentially explain MGV-like mechanisms as approximations to optimal behaviour, MGV cannot explain resource-rational optima as emergent from its architecture.

Yet this limitation is a consequence of methodological choice rather than oversight. We prioritised preserving psychological detail over deriving mechanisms from first principles. The result is a framework that describes what metacognitive components \textit{might} look like but cannot explain why they should take those forms rather than others. Whether dual-counter FOK approximates optimal stopping behaviour, whether satisficing thresholds should decay exponentially rather than linearly, whether tripartite knowledge organisation is functionally necessary: these questions cannot be answered within MGV but could potentially be addressed through resource-rational analysis. The grounding of MGV in resource-rational principles represents one path toward resolving this limitation.

\subsection{Future Directions}

\paragraph{Grounding MGV in Resource-Rational Principles} The most productive path forward may involve selective integration of the two approaches. Some MGV constructs appear directly amenable to resource-rational analysis. The stopping dynamics formalised in Nelson and Narens' retrieval model parallel the optimal stopping structure of memory recall in the resource-rational framework, where feeling-of-knowing serves as the monitoring signal that enables efficient termination of searches unlikely to succeed \citep{callaway2024optimal}. Flavell's cognitive monitoring, which involves inferring the state of ongoing cognitive processes from imperfect signals, shares the partial observability structure of the belief-MDP formulation of cognitive control \citep{lieder2018rational}. The threshold dynamics that govern when to cease retrieval or when to shift strategies may be derivable from VOC calculations rather than stipulated as free parameters.

Other constructs may serve primarily descriptive functions that complement rather than compete with normative analysis. The tripartite knowledge taxonomy (person, task, and strategy knowledge) provides a vocabulary for the features over which resource-rational agents must learn, but the taxonomy itself is not derived from optimality principles. Similarly, the phenomenological vocabulary of metacognitive experiences (feeling-of-knowing, judgment-of-learning, feeling-of-confidence) names the signals that enable metacognitive control, but resource-rational analysis explains why such signals are useful rather than what they feel like. A complete theory may require both the normative grounding that resource-rational analysis provides and the psychological granularity that MGV preserves.

\paragraph{Meta-Reasoning and Meta-Learning in Language Models.} \citet{griffiths2019doing} identify meta-reasoning and meta-learning as two components of human intelligence that current AI systems lack: meta-reasoning concerns efficient allocation of computational resources within a task, while meta-learning concerns efficient use of data across tasks to accelerate future learning. Resource-rational analysis provides formal frameworks for both (see Appendix~\ref{app:resource-rational}). MGV formalises both components from a psychological rather than normative starting point. For meta-reasoning, the models derived from Flavell's cognitive monitoring and Nelson and Narens' retrieval address how to allocate effort during task performance, specifying when the Monitor should trigger the Generator to produce new reasoning and when the Verifier should accept or reject candidate outputs. For meta-learning, the model derived from Nelson and Narens' acquisition addresses how to distribute study across items for future retrieval, deciding which items merit additional encoding effort given current memory states.

Recent work has begun applying rational metareasoning principles to language models. \citet{desabbata2024rational} developed RaM, a VOC-inspired reward function that balances the utility of chain-of-thought reasoning against its token cost. With RaM, models learn to adapt reasoning length to task difficulty, using substantially fewer tokens on easier problems. This demonstrates that VOC-based training can induce adaptive computation, but it addresses only one dimension of meta-reasoning: \textbf{how long to} \texttt{<think>}. The resource-rational frameworks reviewed in Appendix~\ref{app:resource-rational} reveal additional dimensions, including \textbf{what to} \texttt{<think>} about (as in planning, where the agent selects which nodes to expand), \textbf{how to} \texttt{<think>} (as in strategy selection, where the agent chooses among qualitatively different reasoning modes), and \textbf{how intensely to} \texttt{<think>} (as in cognitive control, where the agent modulates the strength of control signals). MGV provides architectural mechanisms for these decisions. A natural direction for future work is to ground these mechanisms in resource-rational principles, deriving MGV's control dynamics from VOC optimisation rather than stipulating them directly.

\bibliographystyle{plainnat}
\bibliography{colm2025_conference}

\newpage
\section{Flavell's Model of Cognitive Monitoring}
\label{app:flavell}

\citet{flavell1979metacognition} describes metacognition as a dynamic control architecture comprising four interacting components: \textit{metacognitive knowledge}, \textit{metacognitive experience}, \textit{goals} (or tasks), and \textit{actions} (or strategies). Rather than viewing metacognition as merely stored knowledge about cognition, Flavell presents it as a dynamic control system. This system operates through continuous interactions between four elements: what agents know about their cognitive capabilities (metacognitive knowledge), what they feel about their current cognitive state (metacognitive experiences), what they aim to achieve (goals), and how they control their thinking (strategies). Central to \citeauthor{flavell1979metacognition}'s model is the principle of reciprocal interaction amongst these components. Rather than operating as independent modules, they form an integrated system characterised by continuous bidirectional influences: metacognitive knowledge guides both strategy selection and the interpretation of ongoing cognitive experiences; these conscious experiences, in turn, update the knowledge base and prompt strategic adjustments; task-goals determine which aspects of metacognitive knowledge become most salient; and the outcomes of chosen actions provide feedback that shapes both immediate metacognitive experiences and longer-term understanding of effective cognitive approaches. This dynamic interplay positions metacognition as a self-regulating system capable of adaptive control over cognitive processes.

\textbf{Initialisation} Let $\mathcal{T}$ be a task and $\mathcal{G}$ be the associated goal. We establish the initial system state:
\begin{equation}
\mathcal{S}_0 = f(\mathcal{T}, \mathcal{G})
\end{equation}
where $(\mathcal{T},\mathcal{G)}$ is self-imposed or externally-imposed.

While \citet{flavell1979metacognition} treats `goals' and `tasks' as equivalent, we maintain a computational distinction to enhance the model's precision: $\mathcal{T}$ represents the specific cognitive enterprise, whilst $\mathcal{G}$ represents the desired outcome or success criteria. This separation enables clearer analysis of metacognitive processes, such as assessing the cognitive demands of $\mathcal{T}$ relative to $\mathcal{G}$, or specifying which approaches to employ for $\mathcal{T}$ to achieve $\mathcal{G}$. For instance, the same reasoning task ($\mathcal{T}$: logical problem-solving) might require different metacognitive assessments depending on whether the goal is speed ($\mathcal{G}_1$: quick approximation) or accuracy ($\mathcal{G}_2$: verified solution).

\textbf{M-G-V (Information Processing) Cycle} For monitoring cycles $\tau = 0, 1,\ldots,T$:

\textbf{WHILE} $\mathcal{S}_\tau = \text{ACTIVE}$:

\begin{enumerate}
    \item \textbf{MONITOR}: \textit{Monitor cognitive status through retrieval of metacognitive knowledge and assessment of metacognitive experience}.

Knowledge activation operates differently across metacognitive cycles, with initial cycles relying solely on task-goal combinations while ongoing cycles incorporate emerging metacognitive experiences. At $t=0$, the system identifies potentially relevant metacognitive knowledge based exclusively on the task-goal pairing. In subsequent cycles ($\tau>0$), the knowledge base expands as metacognitive experiences ($\mathcal{ME}$) from the previous cycle $\tau-1$ triggering additional relevant knowledge. 

\begin{equation}
\mathcal{MK}_\tau = \begin{cases}
\text{retrieve}(\mathcal{MK}, \mathcal{T}, \mathcal{G}) & \text{if } \tau = 0 \\
\mathcal{MK}_{\tau-1} \cup \text{retrieve}(\mathcal{MK}, \mathcal{ME}_{\tau-1}) & \text{if } \tau > 0
\end{cases}
\end{equation}

According to \citet{flavell1979metacognition}, metacognitive knowledge comprises three major categories:
  
  \begin{itemize}
      \item \textbf{Agent Variables} ($\mathcal{MK}_{\text{Agent}}$): Knowledge about cognitive agents' characteristics and capabilities that applies across different cognitive endeavours. These are fundamentally subjective beliefs about processing preferences, strengths, and limitations rather than objective assessments. For computational agents, these may represent \textit{learned self-models} -- representations of performance patterns, processing preferences, and comparative capabilities derived from experience across cognitive tasks.
      
      \item \textbf{Task Variables} ($\mathcal{MK}_{\text{Task}}$): Knowledge about cognitive situation assessment, including: (1) information characteristics (e.g., familiarity, complexity, organisation), and (2) task demands and goals. This knowledge is evaluative -- understanding what task characteristics mean for cognitive processes and goal achievement, not merely recognising the characteristics themselves.
      
      \item \textbf{Strategy Variables} ($\mathcal{MK}_{\text{Strategy}}$): Knowledge concerning the effectiveness of cognitive strategies ($\mathcal{CS}$) and metacognitive strategies ($\mathcal{MS}$). Across different goals and task types, $\mathcal{CS}$ are cognitive operations that address problem-solving procedures such as applying domain-specific algorithms or step-by-step problem decomposition, whereas $\mathcal{MS}$ monitor and regulate such cognitive processes. For instance, chain-of-thought reasoning represents a $\mathcal{CS}$ for solving problems systematically, while deciding to \textit{employ} chain-of-thought based on problem complexity assessment represents a $\mathcal{MS}$. \citet{flavell1979metacognition} explicitly incorporates both strategy types within this category, reflecting his theoretical position that strategy selection constitutes a fundamentally metacognitive process requiring knowledge about when, how, and why particular approaches prove effective under specific conditions.
  \end{itemize}

These categories function as an integrated system: task variables diagnose cognitive demands, strategy variables prescribe responses, and agent variables contextualise both within the agent's capabilities\footnote{The distinction between $\mathcal{MK}_{\text{Strategy}}$ and $\mathcal{MK}_{\text{Task}}$ emerges from their functional roles. $\mathcal{MK}_{\text{Task}}$ enables diagnosis by identifying what makes cognitive enterprises demanding and how task characteristics influence goal achievement probability, whereas $\mathcal{MK}_{\text{Strategy}}$ enables prescription by specifying which cognitive approaches to deploy given those diagnostic assessments. Task variables answer 'what challenges does $\mathcal{T}$ present relative to $\mathcal{G}$?' and strategy variables answer 'which approaches for $\mathcal{T}$ will achieve $\mathcal{G}$?'}.

\citet{flavell1979metacognition} distinguishes between knowledge-based experiences, which ‘are best described as items of metacognitive knowledge that have entered consciousness’ (e.g., suddenly recalling a relevant strategy), and feeling-based experiences, which ‘clearly cannot be described that way’ (e.g., feeling confused).

This dual nature of metacognitive experience -- alternating between immediate phenomenological feelings and knowledge-based assessments -- motivates our formal representation using the exclusive-or operator $\mathcal{\oplus}$. In this formulation, $feel(·)$ captures the pure subjective sensations of cognitive state, while $assess(·)$ represents evaluations informed by metacognitive knowledge. The operator $\mathcal{\oplus}$ thus reflects Flavell’s distinction between feeling-based experiences (phenomenological states that cannot be reduced to knowledge) and knowledge-based experiences (instances of metacognitive knowledge entering consciousness).

Our exclusive-or formalisation captures the observation that these two modes typically alternate rather than blend, though we acknowledge that this binary representation constitutes a modelling simplification of potentially richer interactions. Accordingly, this binary characterisation suggests that, at any given moment, an agent experiences either raw cognitive feelings awaiting interpretation or automatic, knowledge-influenced assessments. The temporal alternation between these exclusive states gives rise to the evolving metacognitive experience that guides subsequent processing.

At this stage, it is notable that \citet[p.~909]{flavell1979metacognition} primarily associates metacognitive experience with the subjective sense of perceived difficulty. Such experiences may involve feeling-of-complexity, comprehension challenges, conceptual opacity, or the sense that material exceeds current capabilities.

\begin{equation}
\mathcal{ME}_{\text{difficulty}}^\tau = 
\begin{cases}
\text{feel}(\mathcal{T}) \oplus \text{assess}(\mathcal{T}, \mathcal{MK}_\tau) & \text{if } \tau = 0 \\
\text{feel}(\mathcal{T}, \text{Outcomes}_{\tau-1}) \oplus \text{assess}(\mathcal{T}, \text{Outcomes}_{\tau-1}, \mathcal{MK}_\tau) & \text{if } \tau > 0
\end{cases}
\end{equation}

Accordingly, $\mathcal{ME}_{\text{difficulty}}^\tau$ evolves through iterative cycle-dependent assessments, progressing from initial coarse-grained feelings to increasingly nuanced evaluations that identify specific challenge sources and their implications for strategy selection. These experiences could help identify specific sources of obstacles and serve to guide the agent’s attentional and regulatory focus.
    
    \item \textbf{GENERATE}: \textit{Control cognitive activity through strategy selection and execution}.

\citet[p.~909]{flavell1979metacognition} emphasises that this stage centres on the selection of cognitive strategies ($\mathcal{CS}_\tau$) through the integration of metacognitive experiences and knowledge. Metacognitive experiences of difficulty ($\mathcal{ME}_{\text{difficulty}}$), whether feeling-based or knowledge-based, function as computational \textit{signals} that indicate cognitive status. However, these signals require interpretation through metacognitive knowledge to guide effective strategy selection. As \citet[p.~906]{flavell1979metacognition} establishes, effective cognitive regulation emerges only when metacognitive experiences combine with metacognitive knowledge, transforming ambiguous feelings into actionable strategic decisions.

The strategy selection process draws upon $\mathcal{MK}_{\text{Strategy}}$, which encompasses knowledge about both metacognitive strategies ($\mathcal{MS}$) and cognitive strategies ($\mathcal{CS}$). Although \citeauthor{flavell1979metacognition} does not specify the exact selection mechanism, his examples suggest a \textit{two-phase pattern-matching} process. In the first phase, $\mathcal{MK}_{\text{Strategy}}$ guides the interpretation of $\mathcal{ME}_{\text{difficulty}}^\tau$, transforming general difficulty signals into precise diagnostic patterns. For instance, metacognitive knowledge might specify ``when experiencing content uncertainty, identify specific unknown terms'' or ``when procedurally confused, assess whether confusion stems from missing steps versus unclear sequence''. In the second phase, these refined difficulty patterns activate corresponding cognitive strategies from $\mathcal{MK}_{\text{Strategy}}$ -- content uncertainty with identified terms triggers seeking definitions, whilst procedural confusion from missing steps activates searching for worked examples. Throughout this process, the selection mechanism integrates agent capabilities ($\mathcal{MK}_{\text{Agent}}$) and task characteristics ($\mathcal{MK}_{\text{Task}}$) to identify the most appropriate strategy for achieving $\mathcal{G}$ given $\mathcal{T}$.

\begin{equation}
\mathcal{CS}_\tau = \text{select}(s \in \mathcal{MK}_{\text{Strategy}} \mid \mathcal{ME}_{\text{difficulty}}^\tau, \mathcal{MK}_\tau, \mathcal{T}, \mathcal{G})
\end{equation}

The selected cognitive strategy is implemented to produce cognitive outcomes ($\mathcal{CO}_\tau$), generating a feedback that is rich in nature, as as it encompasses not only task progress information but also a new context for the next cycle's monitoring and potential strategy adjustment. 

\begin{equation}
\mathcal{CO}_\tau = \text{execute}(\mathcal{CS}_\tau, \mathcal{T}, \mathcal{G})
\end{equation}
    
    \item \textbf{VERIFY}: \textit{Evaluate progress and determine continuation}.
    
Following strategy execution, \citet[p.~909]{flavell1979metacognition} comments that the outcomes potentially `trigger additional metacognitive experiences about how the endeavour is faring'. These evaluative experiences ($\mathcal{ME}_{\text{evaluative}}^\tau$) are about performance rather than difficulty. 

\begin{equation}
\mathcal{ME}_{\text{evaluative}}^\tau = \text{feel}(\mathcal{\mathcal{CO}_\tau}) \oplus \text{assess}(\mathcal{CO}_\tau, \mathcal{MK}_\tau)
\end{equation}

These experiences, again informed and guided by pertinent metacognitive knowledge, instigate the metacognitive strategy of surveying `all that [the agent has] learned to see if it fits together into a coherent whole, if it seems plausible and consistent with [the agent's] prior knowledge and expectations, and if it provides an avenue to the goal' \cite[p.~909]{flavell1979metacognition}.

\begin{equation}
\mathcal{MS}_\tau = \text{select}(s \in \mathcal{MK}_{\text{Strategy}}^{\text{meta}} \mid \mathcal{ME}_{\text{evaluative}}^\tau, \mathcal{MK}_\tau, \mathcal{CO}_\tau, \mathcal{G})
\end{equation}

where 

\begin{equation}
\mathcal{MS}_\tau = \begin{cases}
\text{coherence} & \text{if } \mathcal{ME}_{\text{evaluative}}^\tau \text{ signals fragmented understanding} \\
\text{plausibility} & \text{if } \mathcal{ME}_{\text{evaluative}}^\tau \text{ signals doubtful results} \\
\text{consistency} & \text{if } \mathcal{ME}_{\text{evaluative}}^\tau \text{ signals unexpected outcomes} \\
\text{goal-conduciveness} & \text{if } \mathcal{ME}_{\text{evaluative}}^\tau \text{ signals uncertain progress}
\end{cases}
\end{equation}

$\mathcal{ME}_{\text{evaluative}}^\tau$ signals the need for assessment. For example, ``feeling uncertain about validity of the processed outcome'' or ``sensing incomplete understanding despite completion of process''. These evaluative experiences activate relevant metacognitive strategies from $\mathcal{MK}_{\text{Strategy}}$. For instance, uncertainty about validity triggers plausibility checking, while sensing incompleteness activates coherence assessment to identify gaps. 

$\mathcal{MS}_\tau$ represents the strategic choice to conduct comprehensive evaluation along four possible dimensions: coherence (``do the outcomes form a consistent understanding?''), plausibility (``are the results believable given prior knowledge?''), consistency (``do outcomes align with initial expectations?''), and goal-conduciveness (``do current results provide a pathway to goal achievement?''). The execution systematically evaluates $\mathcal{CO}_\tau$ against relevant knowledge:

\begin{equation}
\mathcal{MO}_\tau = \text{execute}(\mathcal{MS}_\tau, \mathcal{CO}_\tau, \mathcal{MK}_\tau, \mathcal{G})
\end{equation}

\citeauthor{flavell1979metacognition} emphasises that metacognitive experiences can `add to', `delete from', or `revise' the metacognitive knowledge base through Piagetian \citep{flavell1963developmental} mechanisms. The agent observes relationships among goals, strategies, experiences, and outcomes across the complete monitoring cycle.

Let $\Phi_\tau$ represents the complete experience tuple, where $\mathcal{ME}_\tau = (\mathcal{ME}_{\text{difficulty}}^\tau, \mathcal{ME}_{\text{evaluative}}^\tau)$, $\text{Strategy}_\tau = (\mathcal{CS}_\tau, \mathcal{MS}_\tau)$ and $\text{Outcome}_\tau = (\mathcal{CO}_\tau, \mathcal{MO}_\tau)$: 

\begin{gather}
\Phi_\tau = (\mathcal{ME}_\tau, \text{Strategy}_\tau, \text{Outcome}_\tau) \\
\mathcal{MK} = \text{update}(\mathcal{MK}, \Phi_\tau)
\end{gather}

Based on the comprehensive metacognitive evaluation, the system determines its next state:

\begin{equation}
\mathcal{S}_{\tau+1} = \begin{cases}
\text{ACTIVE} & \text{if } \neg\text{goal\_achieved}(\mathcal{CO}_\tau, \mathcal{G}) \\
\text{TERMINATE} & \text{if } \text{goal\_achieved}(\mathcal{CO}_\tau, \mathcal{G}) 
\end{cases}
\end{equation}

\end{enumerate}

\newpage
\section{Nelson and Narens' Model of Metamemory}
\label{app:nelson-narens}

\citet{nelson1990metamemory} theorise metacognitive systems with particular focus on metamemory in the context of self-directed, self-paced learning and retrieval tasks. Their framework establishes metacognition as fundamentally hierarchical, distinguishing between cognitive processes that operate on mental content (object-level) and those that operate on cognitive processes themselves (meta-level). This two-level architecture provides the theoretical foundation for understanding how cognitive systems achieve self-regulation and control during learning activities.

According to their model, the meta-level maintains a dynamic internal representation of the object-level, functioning as a mental simulation that enables the system to monitor current cognitive states and guide transitions towards desired goals. The interaction between levels operates through two distinct information flows, \textit{control} (meta-level $\rightarrow$ object-level) and \textit{monitoring} (object-level $\rightarrow$ meta-level). Control processes enable the meta-level to modify object-level states or processes -- such as allocating study time to difficult material or switching from rote memorisation to elaborative rehearsal strategies. Monitoring processes provide the meta-level with information about current object-level states, updating its internal model of the cognitive situation. These relationships connotes two notable properties: they are logically independent (control does not inherently generate feedback about its effects) and asymmetric (the meta-level maintains a model of the object-level whilst the object-level operates without any corresponding representation of the meta-level). 

\subsection{Acquisition Process}

\textbf{Initialisation} Given a task ($\mathcal{T}$) and goal ($\mathcal{G}$) with a target performance level ($\rho^*$), the agent establishes the \textit{norm of study} ($\mathcal{N}_s$):

\begin{gather}
\mathcal{MK}_0^{\text{STM}} = \text{retrieve}_\theta(\mathcal{MK}, \mathcal{T}, \mathcal{G}) \\
\delta_{\text{retention}} = \text{formulate}(\mathcal{MK}_0^{\text{STM}}, \tau_{\text{delay}}, \mathcal{T}, \mathcal{G}) \\
\mathcal{N}_s = \rho^* \times (1 + \delta_{\text{retention}})
\end{gather}

At the initialisation stage ($\tau = 0$), a global metacognitive parameter ($\mathcal{N}_s$) operationalises abstract goals into quantified mastery criteria, which \citet[p. 130]{nelson1990metamemory} define as `the overall degree of mastery the person believes should be attained during acquisition'. 

Following \citet{ericsson1984protocol}, monitoring operations occur within working memory (STM), with $\mathcal{MK}_0^{\text{STM}}$ denoting the metacognitive knowledge retrieved into this workspace at $\tau = 0$. Information from long-term memory (LTM) may be accessed by first copying it into STM with probability $\theta$ \citep{atkinson1968human}, captured through the notation $\text{retrieve}_\theta(\cdot)$ for this probabilistic access during metacognitive monitoring. The term $\delta_{\text{retention}}$ represents the agent's theory of retention -- beliefs about memory decay over the interval $\tau_{\text{delay}}$.

This formulation reflects Nelson and Narens' insight that effective learning requires anticipatory compensation for memory decay. The model predicts systematic variation in norm-setting behaviour across agents and contexts. For instance, an agent targeting 90\% test performance ($\rho^* = 0.9$) who expects 20\% decay ($\delta_{\text{retention}} = 0.2$) must achieve 108\% mastery during acquisition. Moreover, the framework anticipates differential standards across learning contexts: conceptual understanding tasks ($G_1$, with $\delta_{\text{retention}} = 0.1$) vs. verbatim recall tasks ($G_2$, with $\delta_{\text{retention}} = 0.2$) yield distinct acquisition targets (99\% vs. 108\% respectively) even under identical performance goals.

\textbf{M-G-V (Learning) Cycle} For learning cycles $\tau \in \{1, \ldots, T_\text{learn}\}$, let $\mathcal{T}_\tau = \{i_j : j \in \mathcal{J}_\tau\}$ denote the set of items remaining in the task at cycle $\tau$, where $\mathcal{J}_\tau \subseteq \{1, 2, \ldots, N\}$ represents the indices of items still requiring learning. $\Phi_{\tau}^{\text{STM}}$ represent the cumulative learning experience in working memory. 

\textbf{WHILE} $\mathcal{S}_\tau = \text{ACTIVE}$:

\begin{enumerate}
    \item \textbf{MONITOR}: \textit{Assess current mastery for each item $i_j \in \mathcal{T}_\tau$}.

Monitoring involves retrieving metacognitive knowledge and generating metacognitive experiences about the current learning state.

\begin{gather}
\mathcal{MK}_\tau^{\text{STM}} = \mathcal{MK}_{\tau-1}^{\text{STM}} \cup \text{retrieve}_\theta(\mathcal{MK}, \mathcal{ME}_{\tau-1}) \quad \text{if } \tau > 0 \\
\mathcal{ME}_{\tau,j} =
\begin{cases}
[\text{EOL}_{\tau,j}, \text{null}] & \text{if } \tau = 0 \\
[\text{FOK}_{\tau,j}, \text{null}] & \text{if } \tau > 0
\end{cases}
\end{gather}

where:

\begin{gather}
\text{EOL}_{\tau,j} = \text{feel}(i_j) \oplus \text{assess}(i_j, \mathcal{MK}_\tau^{\text{STM}}) \quad \text{if } \tau = 0\\
\text{FOK}_{\tau,j} = \text{feel}(i_j, \mathcal{CO}_{\tau-1,j}) \oplus \text{assess}(i_j, \mathcal{CO}_{\tau-1,j}, \mathcal{MK}_\tau^{\text{STM}}) \quad \text{if } \tau > 0
\end{gather}

Metacognitive experiences are represented as vectors, reflecting Nelson and Narens' proposal of their multidimensional nature. Both ease-of-learning (EOL) and feeling-of-knowing (FOK) are immediate phenomenological experiences that emerge during cognitive tasks, illustrating how subjective feelings support monitoring functions, serving as the primary input for subsequent control decisions \citep[p. 160]{nelson1990metamemory}. 

% \footnote{For the present learning framework, we represent FOK as scalar value, though \citeauthor{nelson1990metamemory} suggest dual-counter mechanism (where FOK comprises both affirmative and negative components, $\text{FOK}_{\tau,j}^{+}$ and $\text{FOK}_{\tau,j}^{-}$, which will be elaborated in the Retrieval Process section).}

    \item \textbf{GENERATE}: \textit{Transforms monitoring outputs into executable learning actions}. 
    
Resources are allocated inversely proportional to their EOL or FOK, and strategy selection integrates metacognitive inputs to map learning methods to individual items. 

\begin{gather}
r_{\tau,j} = R_{\text{total}} \times \frac{w_j}{\sum_{k=1}^{N} w_k}, \quad \text{where } w_j = (\mathcal{ME}_{\tau,j}[1])^{-1} \\ 
\sigma_{\tau,j} = \text{select}(s \in \mathcal{MK}_{\text{Strategy}} \mid i_j, r_{\tau,j}, \mathcal{ME}_{\tau,j}, \mathcal{MK}_\tau)
\end{gather}

The learning plan $\mathcal{P}_{\tau,j} = (i_j, r_{\tau,j}, \sigma_{\tau,j})$ is executed to produce cognitive outcomes (new memory state). 

\begin{equation}
\mathcal{CO}_{\tau,j} = \text{execute}(\mathcal{P}_{\tau,j})
\end{equation}

    \item \textbf{VERIFY}: \textit{Assess learning progress and determines cycle continuation}. 
    
Judgement-of-learning (JOL) evaluate current mastery levels following cognitive outcomes. 
\begin{gather}
\text{JOL}_{\tau,j} = \text{feel}(i_j, \mathcal{CO}_{\tau,j}) \oplus \text{assess}(i_j, \mathcal{CO}_{\tau,j}, \mathcal{MK}_\tau) \\ 
\mathcal{ME}_{\tau,j} = [\mathcal{ME}_{\tau,j}[1], \text{JOL}_{\tau,j}] 
\end{gather}

For each item $i_j \in \mathcal{T}_\tau$, the agent computes the mastery discrepancy between the current JOL and the norm of study $\mathcal{N}_s$. Items that have reached the mastery criterion are removed from further consideration, so $\mathcal{T}_{\tau+1} = \{i_j : j \in \mathcal{J}_{\tau+1}\}$ contains only items still requiring learning.
\begin{gather}
\Delta_{\tau,j} = \mathcal{N}_s - \text{JOL}_{\tau,j} \\
\mathcal{J}_{\tau+1} = \{j \in \mathcal{J}_\tau : \Delta_{\tau,j} > 0\}
\end{gather}

The metacognitive experiences from each item, together with the strategies used and outcomes obtained, are then packaged into experience tuples and aggregated across all items processed in the current cycle.
\begin{gather}
\Phi_{\tau,j}^{\text{STM}} = (\mathcal{ME}_{\tau,j}, i_j, r_{\tau,j}, \sigma_{\tau,j}, \mathcal{CO}_{\tau,j}) \\
\Phi_{\tau}^{\text{STM}} = \Phi_{\tau-1}^{\text{STM}} \cup \{\Phi_{\tau,j}^{\text{STM}} : j \in \mathcal{J}_\tau\}
\end{gather}

Learning continues as long as any item remains below the mastery threshold:
\begin{equation}
    \mathcal{S}_{\tau+1} = 
    \begin{cases}
        \text{ACTIVE} & \text{if } \mathcal{J}_{\tau+1} \neq \emptyset \\
        \text{TERMINATE} & \text{otherwise}
    \end{cases}
\end{equation}

\end{enumerate}

\subsection{Retrieval Process}

\textbf{Initialisation} Given a retrieval query $\mathcal{Q}$, the agent establishes retrieval goals and accesses contextually relevant metacognitive knowledge for search control.

\begin{equation}
\mathcal{MK}_0^{\text{STM}} = \text{retrieve}_\theta(\mathcal{MK}, \mathcal{Q})
\end{equation}

\citet{nelson1990metamemory} conceptualise FOK through the dual-counter hypothesis: one component accumulates evidence for information presence in memory (affirmative FOK, $\text{FOK}_{\tau}^{+}$), while the other accumulates evidence for information absence, consistent with `knowing not' \citep{kolers1976knowing} (negative FOK, $\text{FOK}_{\tau}^{-}$). This dual mechanism enables both continued search when positive evidence accumulates and efficient termination when negative evidence dominates, preventing exhaustive search behaviour.

The initial thresholds $\lambda_{\text{confidence}}^{(0)}$ and $\lambda_{\text{FOK}}^{(0)}$ are established through the agent's privileged access to personal metacognitive calibration history within $\mathcal{MK}_0^{\text{STM}}$:

\begin{gather}
\lambda_{\text{FOK}}^{(0)} = \text{median}(\{||\text{FOK}|| : \text{successful retrievals in } \mathcal{MK}_0^{\text{STM}}\}) \\
\lambda_{\text{confidence}}^{(0)} = \text{median}(\{\text{confidence}_{\tau} : \text{correct outputs in } \mathcal{MK}_0^{\text{STM}}\})
\end{gather}

FOK thresholds are calibrated based on successful retrievals -- episodes where dual-counter FOK assessment correctly predicted retrieval outcomes, with $||\text{FOK}_{\tau}||$ (L1 norm) capturing the magnitude of metacognitive evidence. Confidence thresholds follow analogous calibration, reflecting historical accuracy at different confidence levels. This personalised approach embodies the No-Magic Hypothesis by utilising recallable metacognitive knowledge whilst accommodating domain-specific variations in metamemory accuracy.

\textbf{M-G-V (Search) Process} For search cycles $\tau \in \{0, 1, \ldots, T_{\text{search}}\}$, let $\mathcal{A}_\tau$ represent the current answer state (retrieved answer or null), and $\Omega_\tau^{\text{STM}}$ represent the cumulative retrieval experience in working memory.

\textbf{WHILST} search is active:

\begin{enumerate}
    \item \textbf{MONITOR}: \textit{Assess Feeling-of-Knowing (FOK) and retrieval accessibility}.

The metacognitive decision to initiate search relies on rapid, preliminary FOK judgement that operates faster than actual recall, enabling efficient search control \citep{reder1987strategy, reder1988strategic}. Following the No-Magic Hypothesis, FOK monitoring accesses recallable item attributes -- acquisition history, partial cues, contextual associations -- rather than directly tapping unconscious memory states.

\begin{gather}
\mathcal{MK}_\tau^{\text{STM}} = \begin{cases}
\text{retrieve}_\theta(\mathcal{MK}, \mathcal{Q}) & \text{if } \tau = 0 \\
\mathcal{MK}_{\tau-1}^{\text{STM}} \cup \text{retrieve}_\theta(\mathcal{MK}, \text{FOK}_{\tau-1}) & \text{if } \tau > 0
\end{cases} \\
\text{FOK}_{\tau} = \begin{bmatrix} \text{FOK}_{\tau}^{+} \\ \text{FOK}_{\tau}^{-} \end{bmatrix} = \begin{cases}
\text{feel}(\mathcal{Q}) \oplus \text{assess}(\mathcal{Q}, \mathcal{MK}_{\tau}^{\text{STM}}) & \text{if } \tau = 0 \\
\text{feel}(\mathcal{Q}, \mathcal{A}_{\tau-1}) \oplus \text{assess}(\mathcal{Q}, \mathcal{A}_{\tau-1}, \mathcal{MK}_{\tau}^{\text{STM}}) & \text{if } \tau > 0
\end{cases}
\end{gather}

At $\tau = 0$, preliminary FOK assessment determines search initiation through rapid accessibility evaluation using the dual-counter system. For subsequent cycles ($\tau > 0$), ongoing FOK monitoring incorporates previous search outcomes ($\mathcal{A}_{\tau-1}$) to reassess continued retrieval likelihood, with both affirmative ($\text{FOK}_{\tau}^{+}$) and negative ($\text{FOK}_{\tau}^{-}$) counters updating based on accumulating evidence.

    \item \textbf{GENERATE}: \textit{Deliberately attend to search cues and execute automatic search}.

Following \citeauthor{nelson1990metamemory}' insight that search execution is automatic once initiated, the generation phase focuses on conscious cue attention rather than strategy selection. The dual-counter FOK hypothesis provides metacognitive control over cue generation intensity, reflecting the principle that monitoring should adaptively influence control processes. 

\begin{equation}
    \mathcal{S}_{\tau} = 
    \begin{cases}
        \text{ACTIVE}_{\text{intensive}} & \text{if } ||\text{FOK}_\tau|| < \lambda_{\text{FOK}}^{(\tau)} \\
        \text{ACTIVE}_{\text{standard}} & \text{if } ||\text{FOK}_\tau|| \geq \lambda_{\text{FOK}}^{(\tau)} \wedge \text{FOK}_{\tau}^{+} > \text{FOK}_{\tau}^{-} \\
        \text{TERMINATE} & \text{if } ||\text{FOK}_\tau|| \geq \lambda_{\text{FOK}}^{(\tau)} \wedge \text{FOK}_{\tau}^{-} > \text{FOK}_{\tau}^{+}
    \end{cases}
\end{equation}

% When FOK magnitude falls below the threshold ($||\text{FOK}_\tau|| < \lambda_{\text{FOK}}^{(\tau)}$), indicating uncertain feeling-of-knowing, the system engages intensive cue attention. This reflects the insight that ambiguous metacognitive signals should trigger more effortful attention. Conversely, when sufficient evidence exists with positive dominance, standard attention suffices, whilst negative dominance triggers termination.

The search intensity logic operates through evidence-based decision making. When the total magnitude of metacognitive evidence falls below the threshold ($||\text{FOK}_\tau|| < \lambda_{\text{FOK}}^{(\tau)}$), insufficient evidence has accumulated from both counters to make a reliable continuation decision. This triggers intensive cue attention to gather additional metacognitive information, preventing premature termination based on weak or ambiguous signals. When sufficient evidence exists ($||\text{FOK}_\tau|| \geq \lambda_{\text{FOK}}^{(\tau)}$), the system evaluates counter dominance: positive dominance ($\text{FOK}_{\tau}^{+} > \text{FOK}_{\tau}^{-}$) indicates sufficient evidence for item presence to warrant continued search with standard attention, while negative dominance ($\text{FOK}_{\tau}^{-} > \text{FOK}_{\tau}^{+}$) provides sufficient evidence for item absence to justify search termination.

If $\mathcal{S}_{\tau} = \text{ACTIVE}$, the agent deliberately attends to retrieval cues that trigger automatic pattern-recognition-guided search, with attention determined by metacognitive confidence. 

\begin{equation}
\text{cue}_\tau = 
\begin{cases}
\text{attend}_{\text{intensive}}(\mathcal{Q}, \mathcal{MK}_{\tau}^{\text{STM}}) & \text{if } \mathcal{S}_{\tau} = \text{ACTIVE}_{\text{intensive}} \\
\text{attend}_{\text{standard}}(\mathcal{Q}, \mathcal{MK}_{\tau}^{\text{STM}}) & \text{if } \mathcal{S}_{\tau} = \text{ACTIVE}_{\text{standard}}
\end{cases}
\end{equation}

Once cues are consciously attended to, the search process $\text{search}_{\text{auto}}(\cdot)$ operates automatically through pattern recognition. Due to this automatic nature, $\mathcal{A}_\tau$ for consecutive cycles $\tau = 0, \ldots, k$ may yield identical results, reflecting the deterministic nature of automatic search. 

\begin{equation}
\mathcal{A}_\tau = \text{search}_{\text{auto}}(\text{cue}_\tau)
\end{equation}

    \item \textbf{VERIFY}: \textit{Evaluate retrieved answers based on confidence, update thresholds, and determine continuation}.

According to \citet{nelson1990metamemory}, confidence governs output decisions for retrieved answers, while FOK governs continuation decisions when no answer is found, with both involving dynamic thresholds that can change during search.

\begin{gather}
\text{confidence}_\tau = 
\begin{cases}
\text{assess}(\mathcal{A}_\tau, \mathcal{Q}, \mathcal{MK}_{\tau}^{\text{STM}}) & \text{if } \mathcal{A}_\tau \neq \text{null} \\
0 & \text{if } \mathcal{A}_\tau = \text{null}
\end{cases} \\
\text{decision}_\tau = 
\begin{cases}
\text{OUTPUT } \mathcal{A}_\tau & \text{if } \mathcal{A}_\tau \neq \text{null} \wedge \text{confidence}_\tau \geq \lambda_{\text{confidence}}^{(\tau)} \\
\text{CONTINUE} & \text{if } \mathcal{A}_\tau \neq \text{null} \wedge \text{confidence}_\tau < \lambda_{\text{confidence}}^{(\tau)} \\ 
\text{CONTINUE} & \text{if } \mathcal{A}_\tau = \text{null} \wedge \text{FOK}_\tau^+ > \text{FOK}_\tau^-\\ 
\text{OUTPUT null (omission)} & \text{if } \mathcal{A}_\tau = \text{null} \wedge \text{FOK}_\tau^- > \text{FOK}_\tau^+
\end{cases}
\end{gather}

This decision structure distinguishes between two primary error pathways identified by Nelson and Narens: (1) \textit{Commission errors} occurring when $\mathcal{A}_\tau \neq \text{null}$ but the outputted answer is incorrect, typically associated with high confidence but incorrect retrieval; and (2) \textit{Omission errors} occurring when search terminates without producing an answer ($\mathcal{A}_\tau = \text{null}$), often following prolonged search with declining FOK. 

The retrieval experience accumulates in working memory, creating a comprehensive search history that informs adaptive threshold adjustment:

\begin{equation}
\Omega_\tau^{\text{STM}} = 
\begin{cases}
[(\text{FOK}_{\tau}, \text{cue}_\tau, \mathcal{A}_\tau, \text{confidence}_\tau)] & \text{if } \tau = 0 \\
\Omega_{\tau-1}^{\text{STM}} \cup [(\text{FOK}_{\tau}, \text{cue}_\tau, \mathcal{A}_\tau, \text{confidence}_\tau)] & \text{if } \tau > 0
\end{cases}
\end{equation}

Following the principle of satisficing \citep{simon1979models}, both confidence and FOK thresholds undergo dynamic adjustment based on accumulated search burden. This reflects the psychological tendency for acceptance criteria to progressively lower as the cost of continued searching increases. The satisficing adjustment factor captures this adaptive mechanism:

\begin{equation}
\beta_\tau = \exp(-\alpha \cdot (\tau + \sum_{(\mathcal{A}_i, \text{conf}_i) \in \Omega_\tau^{\text{STM}}} \mathbf{1}[\mathcal{A}_i = \text{null} \vee \text{confidence}_i < \lambda_{\text{confidence}}^{(i)}]))
\end{equation}

where $\alpha$ represents the satisficing adjustment rate, and the exponential decay function models the psychological burden accumulating from both temporal persistence ($\tau$) and retrieval failures (unsuccessful attempts or low-confidence outcomes). This burden manifests as decreasing acceptance standards, operationalised through threshold reduction:

\begin{gather}
\lambda_{\text{confidence}}^{(\tau+1)} = \lambda_{\text{confidence}}^{(\tau)} \cdot \beta_\tau \\ 
\lambda_{\text{FOK}}^{(\tau+1)} = \lambda_{\text{FOK}}^{(\tau)} \cdot \beta_\tau
\end{gather}

This adaptive mechanism\footnote{Our adaptive threshold adjustment implements \citeauthor{nelson1990metamemory}' `costs/rewards rules' through a simplified model where search burden (costs) drives decreasing acceptance thresholds (reward standards). While this captures the essential cost/reward logic of adaptive satisficing, it abstracts away the multidimensional complexity that separate cost factors (time pressure, cognitive effort) and reward factors (answer importance, confidence benefits) might require.} ensures that answers previously deemed inadequate may become acceptable as search costs accumulate. Consequently, at cycle $\tau+1$, a previously retrieved answer might satisfy the lowered confidence threshold and be output, even though it failed to meet the more stringent earlier criteria.

The search state for the next cycle is determined by:

\begin{equation}
\mathcal{S}_{\tau+1} = 
\begin{cases}
\text{ACTIVE} & \text{if } \text{decision}_\tau = \text{CONTINUE} \\
\text{TERMINATE} & \text{if } \text{decision}_\tau \in \{\text{OUTPUT } \mathcal{A}_\tau, \text{OUTPUT null}\}
\end{cases}
\end{equation}

\end{enumerate}

\newpage
\section{Rational Metareasoning}
\label{app:resource-rational}

The papers reviewed in this section share a common theoretical commitment; each treats a cognitive capacity as a \textit{meta-level decision problem} in which the agent must choose which computations to perform in order to maximise expected reward while minimising computational cost. However, the formal structure of this problem differs across domains.

\paragraph{Contextual bandits} In some domains, each problem is independent. Strategy selection \citep{lieder2017strategy} exemplifies this structure. The agent observes features $\mathbf{f}$ characterising the current problem, selects a meta-level action, and receives feedback. There is no sequential structure within a problem -- only across problems, as the agent learns which actions work well in which contexts. 

\paragraph{Meta-level MDPs} In other domains, the agent faces a sequential decision problem where each meta-level action updates a fully observable state. Planning \citep{callaway2022rational} and memory recall \citep{callaway2024optimal} exemplify this structure. In planning, the state is the decision tree constructed so far -- the agent knows exactly which nodes have been expanded and their values. In memory recall, the state comprises recall progress and elapsed time -- the agent directly observes how much evidence has accumulated toward retrieval.

The ``meta-level'' designation distinguishes these from standard MDPs. A standard MDP formalises interaction with an external environment; a meta-level MDP formalises interaction with an internal, computational environment. States correspond to knowledge states rather than world states, actions correspond to computations rather than physical behaviours, and rewards capture both computational cost and decision quality. Formally, a meta-level MDP is defined as:
\begin{equation}
M = (\mathcal{S}, \mathcal{A}, T, r)
\end{equation}
where $\mathcal{S}$ denotes the set of states, $\mathcal{A}$ the set of meta-level actions, $T(s, a, s')$ the transition function specifying how computations update knowledge, and $r(s, a)$ the meta-level reward.

\paragraph{Belief-MDPs} When the relevant state is only partially observable, the problem becomes a Partially Observable MDP (POMDP). Cognitive control \citep{lieder2018rational} exemplifies this: the control system cannot directly access the full state of perceptual processes, working memory contents, or decision variables, but must infer these from available signals.

Such problems can be reformulated as belief-MDPs, where the state variable encodes what the agent \textit{believes} rather than the true underlying state. Because beliefs must be initialised and updated based on observations, the initial belief state becomes a defining feature of the problem. Formally, a belief-MDP is defined as:
\begin{equation}
M = (\mathcal{B}, b_0, \mathcal{C}, T, r)
\end{equation}
where $\mathcal{B}$ is the set of belief states, $b_0$ is the initial belief state, $\mathcal{C}$ is the set of meta-level actions, $T(b, c, b')$ specifies transition probabilities between belief states, and $r(b, c)$ is the meta-level reward.

\subsection{Domain-Specific Instantiations}

The following sections apply this framework to four domains. Table~\ref{tab:comparison} summarises how each instantiates the relevant formal structure.

\begin{table}[H]
\centering
\begin{tabularx}{\textwidth}{@{}lXXXX@{}}
\toprule
Component & Strategy Selection \citep{lieder2017strategy} & Planning \citep{callaway2022rational} & Cognitive Control \citep{lieder2018rational} & Memory Recall \citep{callaway2024optimal} \\
\midrule
Structure & Contextual bandit & Meta-level MDP & Belief-MDP & Meta-level MDP \\
Context/State & Features $\mathbf{f}$ & State $s$ & Belief state $b$ & State $(t, z_t)$ \\
Observability & Fully observable & Fully observable & Partial & Fully observable \\
Meta-level action & Strategy $s$ & Expansion $a_i$ / $\bot$ & Control signal $c$ & SEARCH / STOP \\
Sequential? & No & Yes & Yes & Yes \\
Learning? & Yes (across problems) & No & Yes (across episodes) & No \\
Approximation & Learned (linear) & Myopic & Learned (LVOC) & Exact (compact state space) \\
\bottomrule
\end{tabularx}
\caption{Structural comparison across domains.}
\label{tab:comparison}
\end{table}

\subsection{Strategy Selection as Rational Metareasoning}
\label{app:strategy}

Multiple cognitive strategies are often applicable to the same problem, yet people's strategy choices depend on task and context, and their adaptiveness increases with experience. \citet{lieder2017strategy} proposed that people learn to approximate a cost-benefit analysis over strategies, predicting how well each will perform on problems with particular features.

\subsubsection{Computational-level analysis}

At the computational level, strategy selection has a distinctive structure that sets it apart from other meta-reasoning problems. Each problem the environment presents constitutes an independent episode. The agent observes a feature vector $\mathbf{f} = (f_1, \ldots, f_n)$ characterising the current problem, selects a cognitive strategy $s$ from a finite set $\mathcal{S}$, and receives feedback in the form of outcome utility $u$ and execution time $t$. Once a strategy is selected, it runs to completion -- there is no sequential structure within a problem. Learning occurs only across problems, as the agent accumulates experience about which strategies work well in which contexts. This structure -- independent problems, observable context, discrete choices -- defines a \textbf{contextual multi-armed bandit}.

The objective is to learn a mapping $m: \mathcal{F} \to \mathcal{S}$ from feature vectors to strategies that maximises expected performance. Intuitively, a good strategy produces high-quality decisions without taking too long. This is formalised as the value of computation:
\begin{equation}
\text{VOC}(s, \mathbf{f}) = \mathbb{E}[U(s(\mathbf{f}); \mathbf{f})] - \gamma \cdot \mathbb{E}[T(s, \mathbf{f})]
\end{equation}
where $U(s(\mathbf{f}); \mathbf{f})$ is the utility of the action selected by strategy $s$, $T(s, \mathbf{f})$ is execution time, and $\gamma$ is the opportunity cost per unit time (e.g., the reward rate). The optimal mapping solves:
\begin{equation}
m^* = \arg\max_m \sum_{\mathbf{f} \in \mathcal{F}} P(\mathbf{f}) \cdot \text{VOC}(m(\mathbf{f}), \mathbf{f})
\end{equation}

However, computing this optimal mapping directly is infeasible because the true VOC of each strategy depends on outcomes revealed only after execution. The agent must learn the mapping from experience. After solving $n$ problems, the agent's history is:
\begin{equation}
h_n = \left((\mathbf{f}^{(1)}, s^{(1)}, u^{(1)}, t^{(1)}), \ldots, (\mathbf{f}^{(n)}, s^{(n)}, u^{(n)}, t^{(n)})\right)
\end{equation}
recording the features, chosen strategy, resulting utility, and execution time for each problem. A learning mechanism $l: \mathcal{H} \to \mathcal{M}$ maps histories to strategy selection mappings, inducing a sequence of improving mappings $m^{(1)}, m^{(2)}, \ldots$ as experience accumulates. The optimal learning mechanism $l^*$ maximises total expected VOC across all problem sequences.

The contextual bandit framing thus highlights two requirements for achieving optimality. First, the agent must generalise from past problems to novel ones using feature-based predictions. Second, the agent must balance exploration (trying uncertain strategies to learn about them) against exploitation (using known-good strategies to maximise immediate reward).

\subsubsection{Model of mental computation}

Having characterised the computational problem, the algorithmic level specifies how the brain might approximate the optimal learning mechanism.

\paragraph{The observability of VOC components} The VOC itself cannot be observed directly -- it is an expectation over future outcomes. However, VOC decomposes into two components that \textit{are} separately observable: when a strategy generates a decision, the resulting utility $u$ and execution time $t$ become available. Recall that:
\begin{equation}
\text{VOC}(s, \mathbf{f}) = \mathbb{E}[U \mid s, \mathbf{f}] - \gamma \cdot \mathbb{E}[T \mid s, \mathbf{f}]
\end{equation}
Because utility and time are observed after each strategy execution, the agent can learn to predict each component independently.

\paragraph{Feature-based function approximation.} The agent learns to predict utility and time from problem features using strategy-specific weights. For each strategy $s$, weights $w_{k,s}^{(U)}$ capture how feature $f_k$ relates to expected utility, and weights $w_{k,s}^{(T)}$ capture how feature $f_k$ relates to expected time. This yields an approximation to the VOC:
\begin{equation}
\widehat{\text{VOC}}(s, \mathbf{f}; \mathbf{w}) = \sum_{k} w_{k,s}^{(U)} f_k - \gamma \cdot \sum_{k} w_{k,s}^{(T)} f_k
\end{equation}
For binary outcomes (success/failure), utility prediction takes a logistic form rather than linear.

\paragraph{Learning the opportunity cost} The agent must also estimate $\gamma$, the opportunity cost per unit time. This represents how much reward the agent foregoes by spending time on the current problem rather than moving to the next. It is modelled as the posterior mean reward rate given cumulative experience: $\gamma = \mathbb{E}[r \mid t_{\text{total}}, r_{\text{total}}]$.

\paragraph{Bayesian weight updating} The weights are learned through Bayesian inference. After each experience $h = (\mathbf{f}, s, u, t)$, the agent updates its posterior distribution over weights. In simple settings, this reduces to Bayesian linear regression for the time model and Bayesian logistic regression for the utility model (see \citet[pp.~66--68]{lieder2017strategy}. The result is a system that learns to predict how well each strategy will perform on problems with particular features.

\subsubsection{Optimal resource allocation}

The optimal policy must balance exploitation of current knowledge against exploration to improve future predictions.

\paragraph{The failure of greedy selection} A greedy policy that always selects the strategy with highest expected VOC ignores the value of learning about uncertain strategies. Suppose the agent has used strategy $s_1$ fifty times with consistent success, yielding a precise estimate of its VOC. Strategy $s_2$ has been tried twice, both times unsuccessfully. A greedy policy would never try $s_2$ again. Yet those failures might have been unlucky, and $s_2$ might actually be superior in certain contexts.

\paragraph{Thompson sampling} The model resolves this dilemma through Thompson sampling. Rather than using point estimates (the posterior means), the agent samples from the full posterior distributions over weights:
\begin{equation}
\tilde{w}_{k,s}^{(U)} \sim P(w_{k,s}^{(U)} \mid h_t), \quad \tilde{w}_{k,s}^{(T)} \sim P(w_{k,s}^{(T)} \mid h_t)
\end{equation}

\paragraph{Optimal policy} Given the sampled weights, the agent computes a sampled VOC for each strategy and selects the strategy that maximises this quantity:
\begin{equation}
s_t = \arg\max_s \widehat{\text{VOC}}(s, \mathbf{f}_t; \tilde{\mathbf{w}}) = \arg\max_s \left[ \sum_k \tilde{w}_{k,s}^{(U)} f_k - \gamma \cdot \sum_k \tilde{w}_{k,s}^{(T)} f_k \right]
\end{equation}
The $\arg\max$ operates over all available strategies $s \in \mathcal{S}$, selecting whichever strategy has the highest predicted value given the current problem features $\mathbf{f}_t$ and the sampled weights $\tilde{\mathbf{w}}$. Crucially, because the weights are sampled rather than fixed at their means, different samples can yield different strategy rankings -- even for the same problem.

This mechanism naturally balances exploration and exploitation through posterior variance. Well-known strategies have narrow posteriors, so samples cluster near expected values -- such strategies are selected only when their expected VOC is genuinely high. Uncertain strategies have wide posteriors whose samples occasionally exceed those of better-known alternatives, prompting the agent to explore them and gather information. As the agent gains experience with a strategy, its posterior narrows, and its selection becomes driven by expected value rather than exploratory variance.

\subsubsection{Evaluation and refinement}

Lieder and Griffiths validated their model against human strategy choices across multiple decision-making paradigms. The model captured context-dependent strategy selection, learning dynamics, and transfer to novel problems. This work contributes to bounded rationality by specifying when people should use which heuristics, framing strategy selection as a learnable metacognitive skill rather than a fixed repertoire.

\subsection{Planning as Rational Metareasoning}
\label{app:planning}

Planning involves mentally simulating future possibilities before acting. Since simulation is costly, the agent faces a resource allocation problem: which aspects of the future should be simulated, and when should deliberation stop? \citet{callaway2022rational} formalised this problem by recognising that planning has a sequential structure absent from strategy selection: each computational action updates the agent's knowledge, and the value of a computation depends on which computations will follow.

\subsubsection{Computational-level analysis}

At the computational level, planning differs fundamentally from strategy selection. In strategy selection, each problem is independent -- selecting a strategy for the current problem does not affect future problems. In planning, the agent performs a sequence of computational actions within a single episode, each of which updates what the agent knows. The value of expanding one branch of a decision tree depends on whether other branches will subsequently be expanded. This sequential dependency structure defines a \textbf{meta-level Markov Decision Process (MDP)}.

The "meta-level" designation distinguishes this from a standard MDP. A standard MDP formalises an agent's interaction with an external environment: states are world states, actions are physical behaviours, and transitions reflect environmental dynamics. A meta-level MDP formalises an agent's interaction with its own internal computational processes: states are knowledge states, actions are computations, and transitions reflect how computations update knowledge. Formally:
\begin{equation}
M = (\mathcal{S}, \mathcal{A}, T, r)
\end{equation}
where $\mathcal{S}$ denotes the set of knowledge states, $\mathcal{A}$ the set of computational actions, $T(s, a, s')$ the transition function specifying how computations update knowledge, and $r(s, a)$ the meta-level reward capturing both computational cost and decision quality.

The objective is to find an optimal policy $\pi^*: \mathcal{S} \to \mathcal{A}$ that maximises total expected meta-level reward -- the quality of the eventual decision minus the cumulative cost of deliberation.

\subsubsection{Model of mental computation}

Having established the meta-level MDP formulation, the algorithmic level specifies how states, actions, transitions, and rewards are concretely represented for the planning problem.

\paragraph{States ($\mathcal{S}$)} A state encodes the agent's current knowledge about the decision problem. Following previous work in rational metareasoning, Callaway et al. represent this knowledge as a decision tree: a directed graph in which nodes represent hypothetical future states and edges represent the actions connecting them. Let $N$ denote the maximum number of nodes in the decision tree. The state is then a vector $\mathbf{s}$ of length $N$, where each element $s_i$ either contains the observed reward at node $i$ or takes the special value $\varnothing$ indicating that node $i$ remains unexpanded. Initially, only the root node (representing the current state) has been expanded with value 0; all other nodes have value $\varnothing$ (e.g., $\mathbf{s}_0 = [0, \varnothing, \varnothing, \ldots]$).

\paragraph{Actions ($\mathcal{A}$)} The action space $\mathcal{A}$ comprises an expansion action $a_i$ for each node $i$ and a termination action $\bot$. Node expansion reveals the reward at node $i$, integrates this value into the total value of the path leading to that node, and adds the node's immediate successors to the search \textit{frontier}. The frontier is defined as the set of nodes eligible for expansion; a node may be expanded only if it belongs to the frontier:
\begin{equation}
\text{frontier}(\mathbf{s}) = \{a_i \mid s_i = \varnothing \land \text{parent}(s_i) \neq \varnothing\}
\end{equation}
When the agent executes the termination action, planning ceases and the agent commits to the plan with highest expected value given the decision tree constructed thus far.

\paragraph{Transition function ($T$)} The meta-level transition function specifies how computational actions modify the state, analogous to how physical actions modify the world state in a standard MDP. Executing expansion action $a_i$ produces a successor state $\mathbf{s}'$ identical to $\mathbf{s}$ except that $s'_i$ is sampled from a node-specific reward distribution $R_i$. Executing the termination action $\bot$ deterministically transitions to a unique terminal state $\mathbf{s}_\bot$. Together, these rules implicitly define a probability distribution $T(\mathbf{s}' \mid \mathbf{s}, a)$ over successor states.

\paragraph{Reward function ($r$)} The meta-level reward function quantifies both the cost of computation and the quality of the resulting decision:
\begin{equation}
r(\mathbf{s}, a) = 
\begin{cases}
\max_{p \in \mathcal{P}} V(\mathbf{s}, p) & \text{if } a = \bot \\
-\lambda & \text{otherwise}
\end{cases}
\end{equation}
where $\lambda > 0$ represents the cost of expanding a single node, $p$ denotes a complete plan (a sequence of nodes from root to leaf), $\mathcal{P}$ is the set of all such plans, and $V(\mathbf{s}, p)$ quantifies the expected value of executing plan $p$ given the current state:
\begin{equation}
V(\mathbf{s}, p) = \sum_{i \in p} 
\begin{cases}
s_i & \text{if } s_i \neq \varnothing \\
\mathbb{E}[R_i] & \text{otherwise}
\end{cases}
\end{equation}
For expanded nodes, the agent uses the observed reward $s_i$; for unexpanded nodes, it uses the prior expectation $\mathbb{E}[R_i]$. The termination reward therefore equals the maximum expected value across all possible plans, reflecting the quality of the decision the agent would make given its current knowledge.

\subsubsection{Optimal resource allocation}

The optimal policy specifies which computational action to take in each state. It is characterised by the optimal Q-function, which quantifies the expected total reward from taking action $a$ in state $\mathbf{s}$ and acting optimally thereafter:
\begin{equation}
Q^*(\mathbf{s}, a) = r(\mathbf{s}, a) + \mathbb{E}_{\mathbf{s}' \sim T(\cdot \mid \mathbf{s}, a)}[V^*(\mathbf{s}')]
\end{equation}
where $V^*(\mathbf{s}) = \max_a Q^*(\mathbf{s}, a)$ is the optimal value function. The optimal policy selects the action with highest Q-value:
\begin{equation}
\pi^*(\mathbf{s}) = \arg\max_a Q^*(\mathbf{s}, a)
\end{equation}

This policy expands a node only when the expected improvement in decision quality exceeds the cost of expansion $\lambda$. Equivalently, it terminates when no expansion has positive expected value of computation.

\paragraph{Intractability} Computing $Q^*$ exactly is intractable for realistically sized decision trees. The state space grows exponentially with the number of nodes, and evaluating each state requires integrating over possible values of unexpanded nodes.

\paragraph{Myopic approximation} Callaway et al. proposed a tractable approximation: at each step, select the computation that would yield the greatest expected improvement if the agent were forced to terminate immediately afterwards. This \textit{one-step lookahead} or \textit{myopic} policy ignores the value of computations that enable future computations, but it preserves the core tradeoff between computational cost and decision quality. Like the optimal policy, it is governed by a single parameter $\lambda$ representing the cost of computation.

This framework does not propose feature-based learning across planning episodes. The agent allocates computation optimally within a single episode; there is no mechanism for transferring knowledge about good planning strategies from one problem to the next.

\subsubsection{Evaluation and refinement}

Callaway et al. evaluated their model using a Mouselab paradigm that externalises planning by requiring participants to click on nodes to reveal values, making the planning process directly observable. Human planning adapted to environmental structure in a manner broadly consistent with the meta-level MDP framework.

However, participants exhibited a pronounced bias towards forward search -- expanding nodes in temporal order -- even when other expansion orders would have been equally adaptive. This suggests that temporally ordered simulation may be a cognitive default, perhaps reflecting adaptation to naturalistic environments where reachable states can only be discovered through forward simulation from the current state.

The authors acknowledge important boundary conditions. The experiments considered only deterministic environments, which ensures that complete planning before acting is optimal; in stochastic environments, planning far ahead may be wasteful as unexpected transitions can render prior deliberation irrelevant. The state spaces were also small and unstructured. Furthermore, the framework provides a computational-level account but not a process-level theory of how people approximate optimal planning. The myopic approximation offers one hypothesis -- that people select computations based on immediate expected value rather than full recursive evaluation -- but this remains to be tested directly.

Despite these limitations, the results suggest that models of efficient resource allocation provide a productive foundation for theories of planning under cognitive constraints.

\subsection{Cognitive Control as Rational Metareasoning}
\label{app:control}

Cognitive control enables the brain to override automatic processes when they conflict with current goals. While exerting control improves performance, it is also effortful. \citet{lieder2018rational} cast cognitive control specification as a sequential decision problem with a crucial additional complexity: unlike planning, where the decision tree is fully observable, the internal state of controlled cognitive systems is only partially observable.

\subsubsection{Computational-level analysis}

At the computational level, cognitive control specification shares the sequential structure of planning -- each control signal affects subsequent options -- but adds the challenge of partial observability. The cognitive control system cannot directly access the full state of perceptual processes, working memory contents, or decision variables. It must infer these internal states from available signals and select control actions based on uncertain beliefs.

When the relevant state is only partially observable, the problem becomes a Partially Observable MDP (POMDP). Such problems can be reformulated as \textbf{belief-MDPs}, where the state variable encodes what the agent \textit{believes} about the environment rather than the true underlying state. Formally:
\begin{equation}
M = (\mathcal{B}, b_0, \mathcal{C}, T, r)
\end{equation}
where $\mathcal{B}$ is the set of belief states (encoding beliefs about both the external environment and internal cognitive state), $b_0$ is the initial belief state, $\mathcal{C}$ is the set of control signals specifying which computations to perform, $T(b, c, b')$ specifies transition probabilities between belief states, and $r(b, c)$ captures outcome utility minus computational cost.

The objective is to find an optimal cognitive control strategy $\pi^*: \mathcal{B} \to \mathcal{C}$ that maximises the expected sum of rewards minus costs. To formalise this, we first define the value function $V_{\pi}(b)$ as the expected cumulative reward from following policy $\pi$ starting in belief state $b$. The optimal value function $V^*(b) = \max_\pi V_\pi(b)$ represents the best achievable expected reward from state $b$.

The total expected value of issuing control signal $c$ in belief state $b$ combines the immediate reward with the expected future value. Because the next belief state is uncertain -- it depends on stochastic factors like stimulus arrivals and internal processing variability -- we write $B_{t+1}$ (a random variable) rather than $b_{t+1}$ (a specific realisation). The \textbf{expected value of control (EVOC)} is then:
\begin{equation}
\text{EVOC}(b, c) = Q^*(b, c) = \mathbb{E}[r(b, c) + V^*(B_{t+1}) \mid B_t = b, C_t = c]
\end{equation}

The optimal strategy selects, in each belief state, the control signal with highest EVOC:
\begin{equation}
\pi^*(b) = \arg\max_{c \in \mathcal{C}} \text{EVOC}(b, c)
\end{equation}

\subsubsection{Model of mental computation}

Having characterised the computational problem, the algorithmic level specifies how the brain might approximate the optimal solution given constraints on time and resources.

\paragraph{The approximation challenge} Computing the EVOC exactly requires estimating the future consequences of control signals across all possible belief state trajectories -- a computation that is itself demanding. Yet cognitive control often operates under severe time pressure: a habitual response may execute within hundreds of milliseconds. The brain cannot afford to solve a complex belief-MDP from scratch on every trial. This tension motivates a learning-based solution: rather than computing the EVOC online, the cognitive control system learns to predict it from experience.

\paragraph{Two obstacles to learning} Learning the optimal Q-function directly faces two obstacles. First, \textit{temporal entanglement}: the value of a control signal depends on which control signals will follow, creating credit assignment problems. Attending carefully at the start of a trial may be valuable or wasteful depending on subsequent attentional adjustments. Second, \textit{dimensionality}: the belief state space encompasses every possible configuration of beliefs about controlled processes -- a space far too large to store separate value estimates for each state.

\paragraph{Feature-based representation} Both obstacles can be addressed by learning a compact, generalising representation. Rather than storing values for individual belief states, the system learns how interpretable features of the situation predict the value of control. Three classes of features are relevant:

\begin{enumerate}
    \item \textbf{State features} $f_k(b)$: aspects of the current context, such as stimulus configuration, task demands, or recent history (e.g., "was the previous target green?")
    \item \textbf{Control signal intensities} $c_l$: the magnitude of each control dimension (e.g., "how strongly to attend")
    \item \textbf{Interaction terms} $f_k(b) \cdot c_l$: how context modulates the value of particular control settings (e.g., "does a recent green target increase the value of attending to green now?")
\end{enumerate}

The \textbf{Learned Value of Control (LVOC)} approximates the EVOC as a weighted sum of these features:
\begin{equation}
\text{LVOC}(b, c; \mathbf{w}) = w_0 + \sum_k w_k^{(f)} f_k(b) + \sum_l w_l^{(c)} c_l + \sum_{k,l} w_{kl}^{(f \times c)} f_k(b) \cdot c_l - \text{cost}(c) - w^{(T)} T
\end{equation}
where $\text{cost}(c)$ reflects the metabolic or opportunity cost of exerting control and $w^{(T)} T$ penalises slow responses.

\paragraph{Bayesian weight updating} The weight vector $\mathbf{w}$ is learned through Bayesian inference. After each experience $e_i = (b_i, c_i, r_i, T_i, b_{i+1})$ -- recording the belief state, chosen control signal, obtained reward, response time, and resulting belief state -- the agent updates its posterior:
\begin{equation}
P(\mathbf{w} \mid e_{1:t}) \propto P(\mathbf{w} \mid e_{1:t-1}) \cdot P(e_t \mid \mathbf{w})
\end{equation}
When a single control signal produces a single reward, this reduces to Bayesian linear regression. When rewards are delayed or depend on sequences of control signals, the update substitutes predicted future values for unobserved outcomes.

\paragraph{Options} The framework extends from elementary control signals to compound cognitive strategies. A single control signal might modulate one parameter (e.g., "attend to the left stimulus"), but complex behaviour often requires coordinating multiple signals simultaneously (e.g., increasing attention while also raising the decision threshold) or executing an entire sequence of operations (e.g., following a multi-step planning routine).

The rational metareasoning framework accommodates such cases by treating cognitive strategies as \textit{options}. An option is a policy combined with an initiation set and a termination condition \citep[pp.~5]{lieder2018rational}; see also \citet{sutton1999between}. That is, an option $o$ can be defined by three components. First, an \textit{initiation set} specifies the belief states in which the option can be started. Second, an \textit{internal policy} specifies what control signals to issue while the option is executing. Third, a \textit{termination condition} specifies when the option finishes.

Once initiated, the option follows its internal policy until the termination condition is met. An elementary control signal can be viewed as an option with trivial structure. A complex strategy is an option with richer structure involving multiple internal steps before termination.

\subsubsection{Optimal resource allocation}

Learning introduces a second-order problem: how should the system select control signals while its value estimates remain uncertain?

\paragraph{The failure of greedy selection.} A greedy policy that always selects the control signal with highest predicted LVOC can fail during learning. Control signals that happen to fail early are abandoned; those that happen to succeed are repeated. The agent never discovers that alternative control signals might be superior in certain contexts.

\paragraph{Thompson sampling} The LVOC model escapes this trap through Thompson sampling. Rather than acting on point estimates (the posterior means), the system samples from its posterior distribution over weights:
\begin{equation}
\tilde{\mathbf{w}} \sim P(\mathbf{w} \mid e_{1:t})
\end{equation}

\paragraph{Optimal policy} Control signals are selected by maximising the LVOC computed with sampled weights:
\begin{equation}
c_t = \arg\max_c \text{LVOC}(b_t, c; \tilde{\mathbf{w}})
\end{equation}

When selecting among options rather than elementary control signals, the policy becomes:
\begin{equation}
\pi^*(b) = \arg\max_{o \in \mathcal{O}} Q^*(b, o)
\end{equation}

As in strategy selection, posterior variance calibrates exploration: uncertain control signals occasionally receive optimistic samples, prompting information gathering.

\subsubsection{Evaluation and refinement}

Lieder et al. validated the LVOC model against five experiments on attentional plasticity and interference control. The model captured how people learn which stimuli to attend based on reward history, how allocation adapts to task difficulty and reward, and how learning transfers to novel stimuli sharing features with previously encountered ones.

The framework rests on several simplifying assumptions. The LVOC approximation assumes that a linear combination of features suffices to capture the value of control, which may not hold in tasks requiring complex, nonlinear interactions among contextual variables. The model also assumes that people can accurately observe their own belief states and the outcomes of their control decisions, yet introspective access to cognitive processes is known to be limited and sometimes distorted. Furthermore, the current formulation treats the feature set as given rather than learned, leaving open the question of how people discover which aspects of a situation are relevant to control.

It is worth clarifying how this framework relates to standard reinforcement learning. Model-based reinforcement learning learns the consequences of acting in the world, such as which locations yield reward following which movements. The LVOC model learns the consequences of thinking in a particular way, such as how strongly attending improves accuracy at the cost of effort. The learning is metacognitive in that its object is not behaviour itself, but the control of the computations that produce behaviour.

Despite these limitations, the LVOC framework offers a principled account of how cognitive control can be both adaptive and learnable. By casting control specification as a belief-MDP and introducing a tractable approximation, the model explains how control exhibits plasticity, adapting through experience to select signals that optimally trade off effort against goal achievement.

\subsection{Memory Recall as Rational Metareasoning}
\label{app:recall}

Retrieving information from memory is not instantaneous. When we cannot immediately recall something, we face a decision: continue searching or give up? Research on metamemory has established that people can judge the likelihood of successful recall -- a capacity termed \textit{feeling-of-knowing} -- and that such judgments influence search duration. \citet{callaway2024optimal} formalised metamemory as a meta-level MDP with the structure of an optimal stopping problem: a higher-order process monitors the progress of a basic recall process and controls how long the search is allowed to continue.

\subsubsection{Computational-level analysis}

At the computational level, memory recall shares the sequential structure of planning -- each moment of continued search is a decision that affects future options. However, in planning, the agent decides both \textit{what} to compute (which node to expand) and \textit{when} to stop; whereas in memory recall, the only decision is whether to continue searching or give up; the computational process itself (evidence accumulation toward recall) proceeds automatically once initiated.

Following classic theories of metamemory \citep{nelson1990metamemory}, the model distinguishes two interrelated processes. The \textit{object-level process} comprises the mechanisms supporting recall itself. The \textit{meta-level process} monitors the rate of progress toward recall and controls how long the search is allowed to continue.

The key insight is that the stopping decision involves a recursive dependency: whether to stop now depends on the expected value of continuing to search, which depends on how long the search would persist, which depends on future stopping decisions. This recursive dependency makes metamemory a sequential decision problem, naturally formalised as a \textbf{meta-level MDP}:
\begin{equation}
M = (\mathcal{S}, s_0, \mathcal{A}, T, r)
\end{equation}
where $\mathcal{S}$ is the set of states encoding current recall progress, $s_0$ is the initial state, $\mathcal{A} = \{\text{SEARCH}, \text{STOP}\}$ comprises two actions, $T$ specifies how continued searching updates the state, and $r$ encodes the benefit of recall and cost of search.

From a rational perspective, one should keep searching as long as the expected utility exceeds the expected cost:
\begin{equation}
a^* = 
\begin{cases}
\text{SEARCH} & \text{if } P(\text{recall}) \cdot U(\text{recall}) > \mathbb{E}[\text{cost}(\text{search})] \\
\text{STOP} & \text{otherwise}
\end{cases}
\end{equation}
The challenge lies in estimating $P(\text{recall})$ and $\mathbb{E}[\text{cost}(\text{search})]$, both of which depend on the unknown strength of the target memory.

\subsubsection{Model of mental computation}

Having characterised the computational problem, the algorithmic level specifies how the object-level and meta-level processes are concretely represented.

\paragraph{Object-level process.} Recall is modelled as a simple evidence accumulation -- a framework widely applied in decision-making and memory research. At each time step $t$, current recall progress $z_t$ is incremented by a noisy sample:
\begin{equation}
z_t = z_{t-1} + x_t \quad \text{where} \quad x_t \sim \mathcal{N}(v, \sigma_x^2)
\end{equation}
The drift rate $v$ captures the strength of the memory: stronger memories accumulate evidence faster. The target is recalled when progress exceeds a threshold $\theta$. Importantly, this threshold is exogenous -- unlike in decision-making models where the agent can choose to commit based on any amount of evidence, the amount of evidence necessary to recall a memory is not under the agent's control (for further discussion on endogenous versus exogenous threshold, see \citet[pp.~18]{callaway2024optimal}.

\paragraph{States} The state $s_t = (t, z_t)$ comprises two quantities: the time elapsed ($t$) and the current recall progress ($z_t$). Together, these provide a complete summary statistic for the entire evidence sequence up to time $t$. The model \textit{assumes} that the meta-level process directly observes this state. This is a simplifying assumption for tractability, not a claim about how people actually monitor their memory -- whether real monitoring tracks underlying memory strength, partial recall progress, or superficial cues remains an open empirical question (see \citet[pp.~4, 19]{callaway2024optimal}).

\paragraph{Inferring memory strength} Although the meta-level process observes recall progress, it does not directly observe the memory's strength $v$. However, observed progress provides approximate information about strength: rapid progress suggests a strong memory; slow progress suggests a weak one. Given observed progress $z_t$ over time $t$, the agent can compute a posterior distribution over memory strength:
\begin{equation}
P(v \mid t, z_t) = \mathcal{N}(v; \mu_t, \sigma_t^2)
\end{equation}
With a weak prior, $\mu_t \approx z_t / t$ -- the average rate of progress. This time-varying belief about memory strength formalises the concept of \textit{feeling-of-knowing}, which is a sense that the target is (or is not) likely to be recalled with further effort.

\paragraph{Transition function} The transition function captures how continued searching updates the state. The next state depends on the increment $x_{t+1}$, which is drawn from $\mathcal{N}(v, \sigma_x^2)$. However, the true memory strength $v$ is unknown to the agent.

To handle this uncertainty, the model \textit{marginalises} over $v$. Marginalisation means integrating out the unknown variable: rather than conditioning on a single value of $v$, the model averages over all possible values, weighting each by its posterior probability $P(v \mid t, z_t)$. This yields a transition probability that reflects the agent's uncertainty about memory strength:
\begin{equation}
T(s_{t+1} \mid s_t, \text{SEARCH}) = \int P(z_{t+1} \mid z_t, v) \, P(v \mid t, z_t) \, dv
\end{equation}
The first term $P(z_{t+1} \mid z_t, v)$ is Gaussian because the increment $x_{t+1} = z_{t+1} - z_t$ is normally distributed with mean $v$ and variance $\sigma_x^2$. The second term $P(v \mid t, z_t)$ is also Gaussian, as derived above. Because the increment can be written as $x_{t+1} = v + \varepsilon$ where $\varepsilon \sim \mathcal{N}(0, \sigma_x^2)$ is noise independent of $v$, the marginal distribution over $x_{t+1}$ is simply the distribution of a sum of two independent Gaussians. This yields a closed-form solution:
\begin{equation}
T(s_{t+1} \mid s_t, \text{SEARCH}) = \mathcal{N}(z_{t+1} - z_t \mid \mu_t, \sigma_x^2 + \sigma_t^2)
\end{equation}
The expected increment equals the posterior mean $\mu_t$ (the agent's current estimate of memory strength), and the variance combines noise in the evidence ($\sigma_x^2$) with uncertainty about the true strength ($\sigma_t^2$). This closed-form solution keeps the model tractable, avoiding the need for numerical integration at each time step.

\paragraph{Reward function} The reward function encodes the benefit of successful recall and the cost of continued search:
\begin{equation}
r(s_t, a) = 
\begin{cases}
U(\text{recall}) & \text{if } z_t \geq \theta \\
-\gamma_{\text{SEARCH}} & \text{if } a = \text{SEARCH} \\
0 & \text{if } a = \text{STOP}
\end{cases}
\end{equation}
where $U(\text{recall})$ is the utility of successful retrieval and $\gamma_{\text{SEARCH}}$ is the explicit (e.g., experimentally imposed cost) and implicit cost (e.g., opportunity cost) per time step of search.

\subsubsection{Optimal resource allocation}

The compact state space $(t, z_t)$ makes exact computation of the optimal policy tractable via backward induction, unlike planning where the exponential state space necessitates approximation.

\paragraph{Optimal value function} The optimal value function can be factorised into two interpretable components:
\begin{equation}
V^*(s_t) = P(\text{recall} \mid s_t) \cdot U(\text{recall}) - \mathbb{E}[t_{\text{stop}} - t \mid s_t] \cdot \gamma_{\text{SEARCH}}
\end{equation}
The first term captures the expected benefit of continued search. $P(\text{recall} \mid s_t)$ is the probability that the target will eventually be recalled given the current state, and $U(\text{recall})$ is the utility of successful retrieval. Their product is the expected utility from recall.

The second term captures the expected cost of continued search. $\mathbb{E}[t_{\text{stop}} - t \mid s_t]$ is the expected number of additional time steps until the search ends, whether by successful recall or by giving up. This quantity depends on both the current state and the agent's future stopping decisions. Multiplying by $\gamma_{\text{SEARCH}}$, the cost per time step, yields the total expected cost of the remaining search. The optimal value is thus the expected benefit minus the expected cost.

\paragraph{Optimal policy} The optimal policy takes the form:
\begin{equation}
\pi^*(s_t) = 
\begin{cases}
\text{SEARCH} & \text{if } \mathbb{E}_{s_{t+1} \sim T(\cdot \mid s_t, \text{SEARCH})}[V^*(s_{t+1})] > \gamma_{\text{SEARCH}} \\
\text{STOP} & \text{otherwise}
\end{cases}
\end{equation}
The expectation $\mathbb{E}_{s_{t+1} \sim T(\cdot \mid s_t, \text{SEARCH})}[V^*(s_{t+1})]$ averages the optimal value of the next state over all possible increments in recall progress, weighted by their probability under the transition function $T$. This expectation represents the expected future value if the agent continues searching for one more time step and then acts optimally thereafter.

The policy thus has a simple interpretation: continue searching if the expected future value exceeds the immediate cost of one more search step; otherwise, stop. This is a one-step lookahead condition, but because $V^*$ already incorporates the consequences of all future optimal decisions, the policy is globally optimal rather than myopic.

\paragraph{Threshold representation} The optimal policy can equivalently be represented as a time-varying threshold on recall progress. If progress $z_t$ falls below this threshold at any moment, the search is terminated. The threshold is not constant because a fixed amount of negative progress provides stronger evidence of low memory strength when generated quickly than when generated slowly. Early in the search, the agent tolerates low progress because uncertainty about memory strength remains high; later, the same progress level may fall below threshold as evidence accumulates that the memory is weak.

\subsubsection{Evaluation and refinement}

The model makes two predictions about the relationship between memory strength and response time. First, stronger memories should be recalled more quickly. This is a straightforward consequence of faster evidence accumulation in the object-level process. Second, stronger memories should be abandoned more slowly. While the meta-level process can quickly identify very weak memories as unlikely to be recalled (their progress falls below threshold early), marginal-strength memories produce ambiguous evidence. It takes longer to accumulate sufficient evidence that the memory is too weak to justify continued search. Together, these predictions explain empirical pattern where response time and judged memory strength are negatively correlated for successful recall (faster responses reflect stronger memories) but positively correlated for unsuccessful recall (longer searches before giving up reflect stronger feeling-of-knowing).

Callaway et al. validated their model using a cued-recall paradigm that allowed objective measurement of memory strength before critical trials. Consistent with the optimal model, participants searched longer before giving up on targets they were more likely to recall, and they prioritised searching for stronger memories when given a choice between targets. These findings support the claim that feeling-of-knowing serves an adaptive function, enabling efficient termination of searches unlikely to succeed while sustaining effort when retrieval remains probable.

The model provides a normative account of how feeling-of-knowing could adaptively guide recall efforts. By formalising metamemory as an MDP, it creates a conceptual link to reinforcement learning: principles governing how people learn to act effectively in the world may also govern how people learn to think effectively in their own minds.

The assumption of direct observation of recall progress $(t, z_t)$ is acknowledged as a simplification. Real metamemory likely involves imperfect monitoring. feeling-of-knowing is known to be influenced by cue familiarity and other factors that may not directly reflect recall progress. Future work might relax this assumption by modelling monitoring as noisy observation of the object-level state, yielding a belief-MDP formulation that more faithfully captures the uncertainty inherent in introspection.

\subsection{Comparing the Value of Computation Across Domains}

Each framework defines a value of computation (VOC) or expected value of control (EVOC) quantifying the benefit of executing a meta-level action. Although the core logic is shared -- weighing computational costs against decision benefits -- the structure of this tradeoff varies. This section compares the formulations along several dimensions.

\subsubsection{Structure of the VOC}

All four formulations share a common structure: VOC equals expected benefit minus expected cost. However, the components differ across domains.

\paragraph{Strategy selection}
\begin{equation}
\text{VOC}(s, \mathbf{f}) = \mathbb{E}[U(s(\mathbf{f}); \mathbf{f})] - \gamma \cdot \mathbb{E}[T(s, \mathbf{f})]
\end{equation}
The benefit is the utility of the decision produced by strategy $s$, while the cost is the opportunity cost of execution time, scaled by the reward rate $\gamma$. A key feature of this formulation is that VOC decomposes into separately observable quantities: utility and time are both revealed after strategy execution, enabling independent learning of each component.

\paragraph{Planning}
\begin{equation}
\text{VOC}(\mathbf{s}, a_i) = \mathbb{E}[\max_p V(\mathbf{s}', p)] - \max_p V(\mathbf{s}, p) - \lambda
\end{equation}
The benefit is the expected improvement in decision quality from expanding node $i$, computed as the difference between expected plan value after expansion and current plan value. The cost is a fixed parameter $\lambda$ per expansion. This formulation focuses on the marginal value of information: each computation is evaluated by how much it improves the agent's ability to identify the best plan.

\paragraph{Cognitive control}
\begin{equation}
\text{EVOC}(b, c) = \mathbb{E}[U(\text{outcome}) - \text{cost}(b, c) + V^*(B_{t+1}) \mid b, c]
\end{equation}
The benefit comprises the utility of the resulting action plus the value of subsequent belief states. The cost includes both immediate effort and a time penalty. A distinctive feature of this formulation is its recursive structure: the value of a control signal depends on which control signals will be issued in the future, as captured by the term $V^*(B_{t+1})$.

\paragraph{Memory recall}
\begin{equation}
V^*(s_t) = P(\text{recall} \mid s_t) \cdot U(\text{recall}) - \mathbb{E}[t_{\text{stop}} - t \mid s_t] \cdot \gamma_{\text{SEARCH}}
\end{equation}
The benefit is the probability of eventual recall multiplied by the utility of successful retrieval. The cost is the expected remaining search time multiplied by the cost per time step. This formulation takes the form of an optimal stopping problem, with a distinctive feature: the threshold for successful recall is exogenous rather than under the agent's control.

\subsubsection{Cost Structure}

\begin{table}[H]
\centering
\begin{tabularx}{\textwidth}{@{}lXX@{}}
\toprule
Domain & Cost formulation & Interpretation \\
\midrule
\textbf{Strategy selection} & $\gamma \cdot T$ & Opportunity cost (learned reward rate $\times$ duration) \\
\textbf{Planning} & $\lambda$ & Fixed cost per expansion \\
\textbf{Cognitive control} & $\text{cost}(b, c) + w^{(T)} \cdot T$ & Effort cost (state/signal-dependent) + time penalty \\
\textbf{Memory recall} & $\gamma_{\text{SEARCH}} \cdot t$ & Fixed cost per time step \\
\bottomrule
\end{tabularx}
\end{table}

Strategy selection and memory recall both frame cost in terms of time, but with different structures: strategy selection \textit{learns} the opportunity cost rate from experience, while memory recall treats it as a fixed parameter.

\subsubsection{Approximation Strategies}

\begin{table}[H]
\centering
\begin{tabularx}{\textwidth}{@{}lXX@{}}
\toprule
Domain & Approximation & Rationale \\
\midrule
\textbf{Strategy selection} & Linear function (learned) & Decomposition enables learning from observable components \\
\textbf{Planning} & Myopic lookahead & Avoids recursive computation; preserves key tradeoff \\
\textbf{Cognitive control} & LVOC (learned) & Compact representation with feature-control interactions \\
\textbf{Memory recall} & Exact (backward induction) & Compact state space makes exact solution tractable \\
\bottomrule
\end{tabularx}
\end{table}

\subsection{Summary}

\begin{table}[H]
\centering
\small
\begin{tabularx}{\textwidth}{@{}lXXXX@{}}
\toprule
 & Strategy Selection & Planning & Cognitive Control & Memory Recall \\
\midrule
\textbf{VOC structure} & $U - \gamma T$ & $\Delta V - \lambda$ & $U - \text{cost} + V'$ & $pU - \gamma t$ \\
\textbf{Formal structure} & Contextual bandit & Meta-level MDP & Belief-MDP & Meta-level MDP \\
\textbf{Sequential} & No & Yes & Yes & Yes (stopping) \\
\textbf{State observable} & Yes & Yes & No & Yes (by assumption) \\
\textbf{Learning} & Yes (across problems) & No & Yes (across episodes) & No \\
\textbf{Cost type} & Opportunity cost & Per-expansion & Effort + time & Per-timestep \\
\textbf{Approximation} & Learned (linear) & Myopic & Learned (LVOC) & Exact \\
\bottomrule
\end{tabularx}
\caption{Summary comparison of VOC formulations.}
\end{table}

Despite these differences, the frameworks share a common normative foundation: \textit{rational metareasoning}. Each asks how a resource-bounded agent should allocate computational effort to maximise expected reward. The VOC provides the currency for this allocation, which is a unified measure of whether a computation is worth performing.

\newpage
\section*{NeurIPS Paper Checklist}

\begin{enumerate}

\item {\bf Claims}
    \item[] Question: Do the main claims made in the abstract and introduction accurately reflect the paper's contributions and scope?
    \item[] Answer: \answerYes{}
    \item[] Justification: The abstract and introduction clearly state this is a theoretical contribution that formalises metacognitive theories, not an empirical study claiming performance improvements.

\item {\bf Limitations}
    \item[] Question: Does the paper discuss the limitations of the work performed by the authors?
    \item[] Answer: \answerYes{}
    \item[] Justification: The paper explicitly states it makes no empirical claims and that translating psychological concepts to computational systems remains speculative.

\item {\bf Theory assumptions and proofs}
    \item[] Question: For each theoretical result, does the paper provide the full set of assumptions and a complete (and correct) proof?
    \item[] Answer: \answerNA{}
    \item[] Justification: This paper formalises existing psychological theories into computational frameworks rather than proving new theoretical results.

\item {\bf Experimental result reproducibility}
    \item[] Question: Does the paper fully disclose all the information needed to reproduce the main experimental results?
    \item[] Answer: \answerNA{}
    \item[] Justification: The paper does not include experiments; it presents theoretical formalisations of cognitive science theories.

\item {\bf Open access to data and code}
    \item[] Question: Does the paper provide open access to the data and code?
    \item[] Answer: \answerNA{}
    \item[] Justification: The paper is purely theoretical and does not involve experiments requiring code or data.

\item {\bf Experimental setting/details}
    \item[] Question: Does the paper specify all the training and test details?
    \item[] Answer: \answerNA{}
    \item[] Justification: The paper does not include experiments.

\item {\bf Experiment statistical significance}
    \item[] Question: Does the paper report error bars suitably and correctly defined?
    \item[] Answer: \answerNA{}
    \item[] Justification: The paper does not include experiments.

\item {\bf Experiments compute resources}
    \item[] Question: For each experiment, does the paper provide sufficient information on the computer resources?
    \item[] Answer: \answerNA{}
    \item[] Justification: The paper does not include experiments.

\item {\bf Code of ethics}
    \item[] Question: Does the research conducted in the paper conform with the NeurIPS Code of Ethics?
    \item[] Answer: \answerYes{}
    \item[] Justification: The theoretical work conforms with ethical guidelines and poses no ethical concerns.

\item {\bf Broader impacts}
    \item[] Question: Does the paper discuss both potential positive societal impacts and negative societal impacts?
    \item[] Answer: \answerNA{}
    \item[] Justification: This is foundational theoretical work without direct societal applications or deployment considerations.

\item {\bf Safeguards}
    \item[] Question: Does the paper describe safeguards for responsible release?
    \item[] Answer: \answerNA{}
    \item[] Justification: The paper poses no risks as it neither releases models nor datasets, only theoretical formalisations.

\item {\bf Licenses for existing assets}
    \item[] Question: Are the creators or original owners of assets properly credited?
    \item[] Answer: \answerNA{}
    \item[] Justification: The paper does not use existing code, data, or model assets.

\item {\bf New assets}
    \item[] Question: Are new assets introduced in the paper well documented?
    \item[] Answer: \answerNA{}
    \item[] Justification: The paper does not release new assets.

\item {\bf Crowdsourcing and research with human subjects}
    \item[] Question: For crowdsourcing experiments and research with human subjects, does the paper include full details?
    \item[] Answer: \answerNA{}
    \item[] Justification: The paper does not involve crowdsourcing or research with human subjects.

\item {\bf Institutional review board (IRB) approvals}
    \item[] Question: Does the paper describe potential risks incurred by study participants?
    \item[] Answer: \answerNA{}
    \item[] Justification: The paper does not involve research with human subjects.

\item {\bf Declaration of LLM usage}
    \item[] Question: Does the paper describe the usage of LLMs if it is an important component of the core methods?
    \item[] Answer: \answerNA{}
    \item[] Justification: LLMs are not used as a component of the core theoretical formalisations presented.

\end{enumerate}

\end{document}